\documentclass{itor}

\usepackage{natbib}%
\usepackage[figuresright]{rotating}
\usepackage{float}

\usepackage{url}
\usepackage{amsmath,amsthm}

\theoremstyle{definition}

\theoremstyle{remark}
\newtheorem{definition}{Definition}

\usepackage[utf8]{inputenc} 
\usepackage[T1]{fontenc}    
\usepackage{booktabs}       
\usepackage{amsfonts}       
\usepackage{microtype}      
\usepackage{xcolor}

\usepackage{tikz}
\usetikzlibrary{shapes.geometric}
\usepackage[noend]{algpseudocode}
\usepackage{adjustbox}
\usepackage{subcaption}
\usepackage{tabularx}
\usepackage{multirow}

\begin{document}

\title{A metaheuristic for crew scheduling in a pickup-and-delivery problem with time windows}

\author{Mauro Lucci\affmark{a,b}, Daniel Sever\'in\affmark{a,b} and Paula Zabala\affmark{a,c,$\ast$}}

\affil{\affmark{a}CONICET, Argentina}
\affil{\affmark{b}Departamento de Matemática (FCEIA), Universidad Nacional de Rosario, Av. Pellegrini 250, Rosario, Argentina}
\affil{\affmark{c}Departamento de Computación (FCEN), Universidad de Buenos Aires, Pabellón 1, Ciudad Universitaria Bs. As., Argentina}

\email{mlucci@fceia.unr.edu.ar [M. Lucci]; daniel@fceia.unr.edu.ar [D. Sever\'in]; pzabala@dc.uba.ar [P. Zabala]}

\thanks{\affmark{$\ast$}Author to whom all correspondence should be addressed (e-mail: pzabala@dc.uba.ar).}

\historydate{Received DD MMMM YYYY; received in revised form DD MMMM YYYY; accepted DD MMMM YYYY}

\begin{abstract}
A \emph{vehicle routing and crew scheduling problem} (VRCSP) consists of simultaneously planning the routes of a fleet of vehicles and
scheduling the crews, where the vehicle--crew correspondence is not fixed through time. 
This allows a greater planning flexibility and a more efficient use of the fleet, but in counterpart, a high synchronisation is demanded.
In this work, we present a VRCSP where pickup-and-delivery requests with time windows have to be fulfilled over a given planning horizon by using trucks and drivers.
Crews can be composed of 1 or 2 drivers and any of them can be relieved in a given set of locations. 
Moreover, 
they are allowed to travel among locations with non-company shuttles, at an additional cost that is minimised.
We tackle this problem with a two-stage sequential approach: a set of truck routes is computed in the first stage and a set of driver routes consistent with the truck routes is obtained in the second one. 
We design and evaluate the performance of a metaheuristic based algorithm for the latter stage.
Our algorithm is mainly a GRASP with a perturbation procedure that allows reusing solutions already found in case the search for new solutions becomes difficult.
This procedure together with other to repair infeasible solutions allow us to find high-quality solutions on instances of 100 requests spread
across 15 cities with a fleet of 12-32 trucks (depending on the planning horizon) in less than an hour.
We also conclude that the possibility of carrying an additional driver leads to a decrease of the cost of external shuttles by about
60\% on average with respect to individual crews and, in some cases, to remove this cost completely.

\end{abstract}


\maketitle

\section{Introduction}

\emph{Vehicle Routing Problems} (VRPs) arise naturally in a wide range of real-world applications.
A well-known particular case is the \emph{Pickup-and-Delivery Problem with Time Windows} (PDPTW) which establishes that customers demand transportation requests from pickup to delivery locations at certain time intervals, see \cite{DUMAS19917}.
However, there are applications where the vehicle routing must be planned together with a crew scheduling \emph{simultaneously}.
Here, crews can change through time, in opposition to use a standard approach where it is considered a fixed vehicle-crew correspondence during the whole planning horizon.
The new approach might improve the productivity since a new driver can continue a task when the previous driver takes a mandatory rest.
However, the whole operation becomes more complex since now separate routes for vehicles and for drivers have to be determined, and both must be synchronised in space and time.
Even worse, the synchronisation leads to the so-called \emph{interdependence} problem: a change in one route may affect the feasibility of other routes, as stated in \cite{Drexl2012}.

In the vast majority of the literature on VRPs, vehicles and crews are treated as inseparable units.
In that case, the working time regulations of the drivers are unnecessarily imposed on vehicles that could be used 24 hours per day. 
Hence, the main contribution of this work is to study how variable crews can be exploited to improve the crew scheduling in a PDPTW that
arose in a long-distance road transport company.
Specifically, we propose a two-stage heuristic algorithm based on an integer programming approach combined with a hybrid metaheuristic that
uses Greedy Randomised Adaptive Search Procedure (\cite{Resende2016}), Variable Neighbourhood Descent (\cite{HANSEN2001449}) and Iterated Local
Search (\cite{ILS}).
Each of the mentioned metaheuristics have been successfully applied to provide good feasible solutions to difficult
combinatorial optimization problems. 

The rest of the paper is structured as follows. In Section \ref{SECTION-DESCRP}, we colloquially describe our problem.
In Section \ref{SECTION-LITERATURE}, we analyse some other related works and the connection between them and ours.
In Section \ref{SECTION-SOLUTION-APPROACH}, we formalise the problem and explain how it is addressed.
In Section \ref{OurAlgorithm}, we present our algorithm.
In Section \ref{SECTION-COMPUTATIONAL-EXPERIMENTS}, we carry out several experiments in order to measure the performance of our algorithm in instances of different sizes as well as to evaluate the features of our particular problem.
Finally, in Section \ref{SECTION-CONCLU}, we draw some conclusions and comment on future works.

\section{Problem description} \label{SECTION-DESCRP}

In our problem, a set of pickup-and-delivery requests has to be fulfilled over a given planning horizon, using the available resources of the company: trucks and drivers.
Each request involves the transportation of an item between two different locations, known as the pickup and delivery locations, with certain temporal constraints.
There is a time window for the pickup and one for the delivery of each request, indicating when the service is allowed to begin.
The time window opens once a day, from an initial day, during a certain time interval, e.g., from the second day of the planning horizon onward and from 10:00am to 6:00pm.
Trucks may arrive earlier, but they must wait until the time window opens to proceed with the service.
The delivery does not have a deadline, but there is a penalty with the delay, where the cost is proportional to the days elapsed since the delivery time window opens for the first time and until the item is delivered.
The duration of the service time is 1 hour, i.e., it takes 1 hour to load or unload an item.

The company's trucks and drivers are homogeneous, except for their initial locations at the beginning of the planning horizon.
They do not have to return to their initial locations at the end of the planning horizon, but the final locations become initial in the next planning.
Trucks have enough capacity to transport any request on a single trip, but just one at a time.
Trucks require drivers when transporting a request and during the service time, and at such circumstances, the drivers are considered to be \emph{working}.
In any other circumstance, drivers are considered to be \emph{resting}.
Rest begins when the driver descends from the vehicle and ends when she/he ascends to that vehicle or another.
There are several factors that regulate rest times, e.g. labour law, collective bargaining agreements, company policy.
The company requires drivers to rest at least 12 hours in every 24-hour interval (this rest might be split) and at least a whole day in any 7-day interval, these constraints will be called $L_1$.
For the last part of this work, we also consider more restrictive variants where there are cumulative constraints limiting the amount of hours worked every week, called $L_2$, and minimum rest duration constraints between consecutive work periods, called $L_3$.


There is a set of locations where the trucks and drivers are allowed to stop, and they are not allowed to stop anywhere else or en route.
This set contains all the pickup and delivery locations and the initial location of the trucks and drivers.
The travel time between the locations is known and constant throughout the day.
Drivers can be relieved at any location and at any time (before, during or after performing a request), even if this causes a detour in the route.
The driver that descends may immediately ascend to another truck or may stay resting (until she/he ascends to another truck).
In our problem, drivers can be transported with shuttles (external to the company) for an additional cost.
In a more restrictive variant of this problem, which we refer as \emph{no-shuttle}, drivers are only allowed to travel by company's trucks.

The main feature of the problem is that the crews can have 1 or 2 drivers.
In the latter case, both are considered to be working 
and each of them can be relieved independently of the other.
This adds great flexibility since a truck could be used as a ``shuttle'' to transport the drivers among the locations.
For example, a driver who descends at a location $l_1$ and who needs to ascend at a location $l_2$ might take advantage of some truck going from $l_1$ to $l_2$ in order to reach $l_2$.
Of course, this is possible as long as the additional driver fits into the intermediate truck (recall that no more than two drivers are allowed per crew) and as long as all these events are synchronised.

The whole operation consists of determining separated routes for the trucks and the drivers, but synchronised in space and time, to fulfill all the requests at minimum operation costs. 
The costs are associated with the total distance travelled by the trucks, the penalties for the delay in the deliveries, and the usage of the external shuttles.

\section{Literature review} \label{SECTION-LITERATURE}

One of the pioneer works that deals with the PDPTW is due to \cite{DUMAS19917}. The authors present an integer programming based
exact algorithm for this problem, which allows multiple vehicles and depots but, on the other hand, no distinction is made between vehicles and drivers, i.e., a fixed vehicle-crew correspondence, nor does it consider drivers' breaks.



The earliest papers where regulations regarding drivers’ working hours are contemplated still consider a fixed vehicle-crew correspondence. Also, changes of drivers on a vehicle route are not allowed, i.e., the entire trip of a vehicle is covered by a single driver.
Therefore, each vehicle trip must satisfy the driver’s work rules, causing that the truck remains ``idle'' while the driver rests.
Within this group \cite{xu2003solving} stands out.
The authors present an exact algorithm for solving a PDPTW proposed by Manugistics (a company that develops software for logistic companies, now part of JDA Software). Their problem considers multiple pickup time windows, multiple delivery time windows, maximum driving time restrictions, and it does not allow a daily rest period to be taken before the maximum daily driving time is exhausted.
They solve the problem with a column generation based approach.

Although the fixed vehicle-crew correspondence is barely abandoned throughout the extensive literature on VRPs for road transport, there are several works that effectively deal with simultaneous Vehicle Routing and Crew Scheduling Problems (VRCSPs).
A fairly complete review (until then) can be consulted in \cite{Drexl2013}, while here we only cite some featured or recent works.

\cite{HOLLIS2006133} is one of the first works where the design of truck routes and scheduling of crews are divided, thus addressing
a simultaneous VRCSP that arises from a mail distribution problem faced by the Australia Post.
In this work, a driver can descend and change to another vehicle upon arrival at a depot.
The authors solve the problem in two stages. First, they determine feasible vehicle routes by solving a path-based MIP model with a heuristic column generation approach. Then, they generate worker and vehicle schedules that fit in the routes obtained in the first stage. 


One of the works that inspired us is \cite{Drexl2013} which addresses a simultaneous VRCSP for long-distance road transport according to the EU legislation. 
In this problem, drivers can only change trucks in relay stations, but breaks and rests can be taken anywhere en route.
Crews are limited to single drivers and there are non-company shuttles to transport the drivers among the stations for an additional cost.
The authors follow a two-stage decomposition of the problem: first, they deal with the vehicles, and after that with the drivers.
A Large Neighbourhood Search method is proposed for both stages, but making some simplifications in order to keep the computational cost low, e.g. drivers are forced to visit some relay station to take a daily rest when a fixed time is expired, and they can only change truck after a daily rest.
These simplifications are probably the reason why the authors conclude that a simultaneous approach does not outperform a fixed vehicle-crew correspondence on an instance coming from a major German freight forwarder with 2800 requests, planning horizon of 6 days, 1975 locations, 1645 trucks and drivers, 43 depots, and 157 additional relay stations.

There are few recent analyses concerning crews with more than one driver in the road transport sector, e.g. in \cite{goel2019team}.
This work investigates under which conditions it is convenient to use single or team driving in European road freight transport, and concludes that the latter is beneficial for longer routes.
However, a fixed vehicle-crew correspondence is assumed, and one driver of the team can rest while the other drives.
Besides, some tasks, such as loading or unloading goods, can be performed in parallel, which reduces times.
The authors implement a Hybrid Genetic Search metaheuristic, which is tested on instances with 100 customers and planning horizon of 6 days. Note that in all these studies, a crew is handled as an indivisible unit, differing from our problem where any member of a crew can be arbitrarily changed at any time and location.

Other scenarios where a synchronisation between vehicles and crews happens are in the airline sector, although, unlike road transport, planes cannot interrupt their flights for the crew to change or rest. In \cite{KASIRZADEH2017111}, a review is
presented about these problems and methodologies are proposed.
\cite{SALAZARGONZALEZ201471} addresses simultaneously the aircraft routing, crew pairing and fleet assignment problems, modeled as a VRP
with multiple depots and changes in the crew among time.
\cite{Lam2015} considers another simultaneous VRCSP with time windows where airplanes are operated by crews that are able to interchange
vehicles at different locations.
Also, the operating time of each crew is limited, although they can travel as passengers outside their working hours.
The authors first propose an integer programming formulation and a constraint programming formulation that model the problem in a single stage.
Then, they consider a two-stage resolution method (similar to \cite{Drexl2013}) where at the first stage, a VRP is formulated as a constraint programming model and then solved via a Large Neighbourhood Search heuristic, and the second one is a set covering problem solved with a standard column-generation approach.
In contrast to \cite{Drexl2013}, the results presented in \cite{Lam2015} show that the ``vehicle interchange'' feature and the fact that the crews can wait in a location for an unlimited amount of time lead to significant better solutions.

On the other hand, in the context of health care transportation, vehicles as well as crews are usually heterogeneous (e.g.~sometimes particular equipment
must be transported with special aircraft; medical staff can be for personal care, rehabilitation services, postoperative care and domestic help services).
In addition, maximum length of routes are limited to daily working hours of employees. See, for instance, \cite{LIN2018151,NASIR2020113361}.

On the subject of public transport, in \cite{Ma2016}, a specific VRCSP is presented to solve a public transport problem according to Chinese regulations.
Here, a set of trips is already assigned to each bus (including the departure and the arrival time of each trip) and a daily bus crew scheduling problem is solved via a Variable Neighbourhood Search. The latter problem takes into account a set of restrictions on driver schedules: the daily working time for a driver is limited to a certain number of hours, a driver is normally required to have a break if the continuous driving time has reached a certain limit, among others limitations. The problem also allows drivers to be relieved at any moment while a bus is parked at a location. Unlike ours, public transport has timetabled trips.

\section{Solution approach}\label{SECTION-SOLUTION-APPROACH}

For a better understanding of our work, let us begin with a preliminary description of how we tackle our problem.
The concepts presented here are further developed in the rest of the paper.

We follow a two-stage decomposition approach.
In the first stage, we route the trucks in order to fulfill all the requests, without worrying about the drivers.
This means that we suppose that the trucks are driven by themselves or by fictional drivers who do not need to rest.

As a result, a set of \emph{tasks} emerges from the truck routes, each of them can be the pickup/delivery of a request or a trip between two locations, and it has a \emph{tentative} start time.
Let us clarify the meaning of tentative.
As far as trucks are concerned, the $i$-th task of the route could be indistinctly scheduled at any time from the end of the $(i-1)$-th task to the beginning of the $(i+1)$-th task, as long as the pickup/delivery time window is still open in the case that the $i$-th task is a pickup/delivery.
However, it is worth mentioning that the start time is not indifferent to the drivers, e.g., a task could be delayed to allow a driver to rest.
Thus, each task has a particular time window of possible start times, which is \emph{dynamic} since it depends on the current truck route (note that changing the start time of a task affects the time window of the contiguous tasks).
This is a major difference with respect to \cite{Drexl2013}, where the time windows of the tasks are computed at once before entering the second stage.

\begin{figure}
    \centering
\includegraphics[scale=0.50]{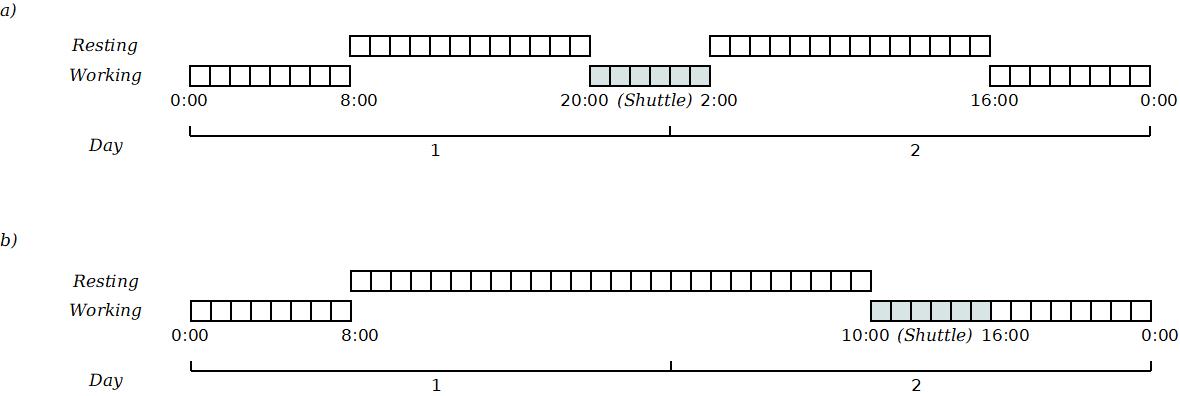}
    \caption{Timelines for a driver who has to take a shuttle.}
    \label{SHUTTLE-EXAMPLE}
\end{figure}

In the second stage, we have to determine the concrete start time and the crew for all the tasks. 
Observe that each task might have a different crew since it is possible to relieve the drivers in any location.
We consider the option of using external \emph{shuttles} (e.g., taxis), so drivers can ascend to a truck in a different location from where they are.
The benefits of using shuttles are clear but, in counterpart, they incur additional costs that should be minimised.
The duration of the shuttles is not negligible in the case of long-distance transport, and since one should consider that the driver is not resting while travelling, the departure time of the shuttles really matters, as the next example illustrates.
Suppose that a driver works the first day from 0:00 to 8:00 and the second day from 16:00 to 0:00 and that she/he needs an intermediate shuttle of 6 hours long, then the shuttle can only depart the first day from 20:00 to 22:00 in order to meet the stipulated rest in every 24-hour interval.
One of such scenario is depicted in Figure \ref{SHUTTLE-EXAMPLE} (a).

Considering that the duration of the shuttles cannot be ignored, and the fact that deciding the best departure time for them involves another optimization problem, we pick a compromise solution where shuttles are always delayed as much as possible.
In the previous example the shuttle would depart the second day at 10:00 and the solution would become infeasible since the driver rests less than 12 hours on the second day, see Figure \ref{SHUTTLE-EXAMPLE} (b).
This decision is arbitrary and a same treatment could be done, for instance, if they are scheduled as early as possible. For the same reason, the possibility of splitting shuttles is not addressed in this work.
Although these simplifications
might compromise the solutions space, perhaps
they are more consistent with real-world applications than \cite{Drexl2013}, where the shuttle travel time is ignored in the working time calculation.
As a final consideration, we assume that there are as many shuttles as required in any location, but they only carry one person at a time. 

The reason behind using a sequential approach relies on keeping the whole operation simpler, as both objects (trucks and drivers) are not dealt at once. 
In addition, the first stage is fairly standard (a PDPTW with no depots and penalties for late deliveries) and can
be resolved by known means.
Our work focuses on the recently presented second stage for our problem.
We propose a metaheuristic based algorithm, which is developed in Section \ref{OurAlgorithm}.

\medskip

The rest of this section is devoted to present the mathematical models for our problem. Let us first present the necessary elements to define it.
\begin{itemize}
    \item $H$ is the number of days of the planning horizon,
    \item $L$ is the set of locations where the trucks and drivers can stop,
    \item $R$ is the set of requests,
    \item $V$ is the set of trucks,
    \item $D$ is the set of drivers,
    \item $l^p_r$, $l^d_r$ are the pickup or delivery location corresponding to $r \in R$, respectively,
    \item $[a^p_r, b^p_r]$, $[a^d_r, b^d_r]$ are the pickup or delivery time window corresponding to $r \in R$, respectively,
    \item $day^p_r$, $day^d_r$ are the initial day where $r \in R$ can be picked up or delivered, respectively,
    \item $c_r$ is the cost per day for delivering the request $r \in R$ late,
    \item $l_v \in L$ is the initial location of each truck $v \in V$,
    \item $l_d \in L$ is the initial location of each driver $d \in D$,
    \item $L_D \doteq \{l_d : d \in D\}$, i.e.~the sets of initial locations of drivers,
    \item $travelTime^T: L \times L \rightarrow \mathbb{R}^+_0$ and $travelTime^S: L \times L \rightarrow \mathbb{R}^+_0$ are functions that return the travel time (in hours) between two locations by truck or by shuttle, respectively,
    \item $travelDist: L \times L \rightarrow \mathbb{R}^+_0$ is the function that returns the distance between two locations, and
    \item $shuttleCost: L \times L \rightarrow \mathbb{R}^+_0$ is the function that returns the cost of a shuttle between two locations.
\end{itemize}

\subsection{First stage: determination of truck routes}

The first stage consists in constructing valid routes for the trucks, such that all the requests are fulfilled during the planing horizon.
A truck route is valid if
it starts from the truck's initial location, 
the time windows are obeyed along the route,
a pickup and the corresponding delivery are performed by the same truck,
each pickup occurs prior to the corresponding delivery,
and the truck does not pick up another request until the current one is delivered.
No driver constraint is considered during this stage. We contemplate two objectives:
\begin{itemize}
    \item $Z_1$: deliver the requests as soon as possible since there is a penalty $c_r$ for a request
    $r$ not delivered during $day^d_r$ proportional to the number of delayed days, i.e.,
    minimise $\sum_{r \in R} c_r (d^d_r - day^d_r)$, where $d^d_r$ is the day $r$ is delivered.
    \item $Z_2$: minimise the total distances travelled by the truck fleet.
\end{itemize}

Although the previous paragraph already describes the first stage 
(and the reader may understand the rest of this work without further details), we give in the Appendix a (standard pick-up and delivery) multi-objective integer linear programming formulation for the sake of completeness.
Besides, this formulation performs well when used to determine the truck routes on small and medium-sized instances.

\subsection{Second stage: determination of driver routes}

Truck routes can be segmented in pickup, delivery, and trip \emph{tasks}.
A pickup/delivery task consists of loading/unloading a request in some location, and its duration is given by the service time (1 hour).
A trip task consists of driving from an origin to a destination location, and its duration is given by the travel time.
For example, suppose that the route of the truck $v$ begins with the transportation of a request $r$.
Then, this route is segmented in the following tasks: a trip task from $l_v$ to $l^p_r$ (whenever $l_v \neq l^p_r$), a pickup task in $l^p_r$, a trip task from $l^p_r$ to $l^d_r$, and a delivery task in $l^d_r$.
Tasks are the smallest pieces of work that must be entirely performed by the same crew.
But a truck can have a different crew for each task since drivers are allowed to change in any location.

Now, let us introduce the following necessary elements:

\begin{itemize}

\item $T^p$, $T^d$, and $T^t$ are respectively the sets of pickup, delivery, and trip tasks involved in the routes constructed in the first stage.

\item $T \doteq T^p \cup T^d \cup T^t$

\item For each $t \in T$, $p_t \in L$ is the origin, $q_t \in L$ is the destination, and $a_t \in \mathbb{R}^+$ is the duration of $t$.

\item $req : T^p \cup T^d \rightarrow R$ is the function that returns the request associated to a given pickup or delivery task.

\end{itemize}

For instance, a trip task $t_1 \in T^t$ from a location $l_1$ to a location $l_2$ of 4 hours long has $p_{t_1} = l_1$, $q_{t_1} = l_2$, and $a_{t_1} = 4$.
A pickup task $t_2 \in T^p$ associated to a request $r$ has $p_{t_2} = q_{t_2} = l^p_r$, $a_{t_2} = 1$, and $req(t_2) = r$; whereas a delivery task $t_3 \in T^d$ associated to $r$ has $p_{t_3} = q_{t_3} = l^d_r$, $a_{t_3} = 1$, and $req(t_3) = r$.
The next definition states when a schedule for the tasks is valid.

\begin{definition}\label{Def Start Time Assigment}
Given $T$ and $req$,
a start time assignment for $T$ is a function $\delta: T \rightarrow \mathbb{R}_0^+$
such that:
1)~whenever two tasks $t_i, t_j \in T$ are consecutive in any truck route then $\delta(t_i) + a_{t_i} \leq \delta(t_j)$,
2)~$\delta(t) \in [a^p_{req(t)}, b^p_{req(t)}]$ for each $t \in T^p$, and
3)~$\delta(t) \in [a^d_{req(t)}, b^d_{req(t)}]$ for each $t \in T^d$.
\end{definition}

Condition (1) imposes \emph{precedence constraints} on tasks of the same truck route: $t_i$ must be finished by the time $t_j$ starts whenever
both tasks are assigned to the same truck and performed consecutively.
Conditions (2) and (3) enforce the pickup and delivery time windows.
While the first stage already assigns a tentative start time for the tasks of $T^p \cup T^d$, the possibility of rescheduling them during the second stage increases the planning flexibility of the crews.
It is worth mentioning that the three conditions guarantee that the truck routes remain valid when the tasks are scheduled according to $\delta$.
The following definition provides a necessary condition for a driver to perform a sequence of tasks.

\begin{definition} \label{defstarttime}
Given $T$, $d \in D$ and a start time assignment $\delta$ for $T$, a driver route for $d$ is a sequence of tasks $\langle  t_0,\ldots,t_k \rangle$ with $t_0,\ldots,t_k \in T$ such that
1) $travelTime^S(l_d,p_{t_0}) \leq \delta(t_0)$, and 
2) $\delta(t_i) + a_{t_i} + travelTime^S(q_{t_i}, p_{t_{i+1}}) \leq \delta(t_{i+1})$ for all $i = 0, \ldots, k-1$.
\end{definition}

Condition (1) ensures that the driver can perform the first task of the route when departing from her/his initial location, and condition (2) determines whether the driver can perform two tasks consecutively; in both cases possible shuttle trips are taken into account.

The second stage consists in deciding a start time assignment for the tasks and constructing valid routes for the drivers, such that each
task is covered by one or two drivers. Below, we elaborate when a driver route is valid, where it can be seen as certain kind of directed
paths in the following digraph.

Given a start time assignment $\delta$ for $T$, consider the simple digraph $G_{\delta} = (\mathcal{V},\mathcal{A})$
with weights $w \in {\mathbb{R}_0^+}^E$. The node set is
$$\mathcal{V} \doteq \{s_l : l \in L_D\} \cup \{n_t : t \in T\} \cup \{\tilde{s}\},$$
where $s_l$ is referred to as \emph{source node} for each initial location $l$ of some driver,
$n_t$ as \emph{task node} for each task $t$ and $\tilde{s}$ as the \emph{sink node}.
The arc set $\mathcal{A}$ has three types of weighted arcs:
\begin{itemize}
    \item From each source node $s_l$ to each task node $n_t$ such that a driver is able to perform $t$ when departing from $l$
    at the beginning of the planning horizon, i.e, $travelTime^S(l,p_t) \leq \delta(t)$.
    The weight of this arc is the cost of a shuttle trip from $l$ to $p_t$,
    i.e, $w(s_l, n_t) = shuttleCost(l, p_t)$.
    \item Between each pair of task nodes $(n_{t_i}, n_{t_j})$ such that a driver is able to perform $t_i$ followed by $t_j$, i.e., $\delta(t_i) + a_{t_i} + travelTime^S(q_{t_i},p_{t_j}) \leq \delta(t_j)$.
    Here, $w(n_{t_i}, n_{t_j}) = shuttleCost(q_{t_i}, p_{t_j})$.
    \item From each source node or task node $u$ to $\tilde{s}$. The weight $w(u,\tilde{s})$ is always 0.
\end{itemize} 

Following the usual terminology, given $a,b \in \mathcal{V}$, a directed walk $P$ from $a$ to $b$ in $G_{\delta}$ is a finite sequence of nodes $u_1, u_2 \ldots,u_{k-1}, u_k$ such that $u_1 = a$, $u_k = b$ and $(u_i,u_{i+1}) \in \mathcal{A}$ for all $1 \leq i \leq k-1$.
When all the vertices visited by $P$ are distinct, we say that $P$ is a \emph{directed path} from $a$ to $b$, also referred to as
$(a,b)$-directed path, and if the only repeated vertices are $v_1$ and $v_k$, then $P$ is called \emph{directed cycle}.
The set of nodes visited by $P$ is denoted with $\mathcal{V}(P)$ and the set of arcs visited by $P$ with $\mathcal{A}(P)$.
The cost $w(P)$ of $P$ is defined as $\sum_{e \in \mathcal{A}(P)} w(e)$.

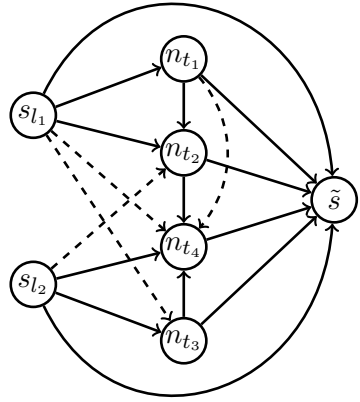
\begin{figure}
\centering
\begin{minipage}{0.4\textwidth}
\begin{tabular}{c|ccccc}
\toprule
 & $p_t$ & $q_t$ & $a_t$ & $t \in$ & $\delta(t)$\\
\midrule
$t_1$ & $l_1$ & $l_1$ & 1 & $T^p$ & 1\\[1pt]
$t_2$ & $l_1$ & $l_2$ & 3 & $T^t$ & 3\\[1pt]
$t_3$ & $l_2$ & $l_2$ & 1 & $T^d$ & 4\\[1pt]
$t_4$ & $l_2$ & $l_2$ & 1 & $T^p$ & 8\\
\bottomrule
\end{tabular}
\end{minipage}
\begin{minipage}{0.35\textwidth}
\begin{tikzpicture}[every node/.style={draw, circle, minimum size=0.6cm, inner sep=0pt, line width=1pt}]

\node (1) at (0,3.75) {$n_{t_1}$};
\node (2) at (0,2.5) {$n_{t_2}$};
\node (4) at (0,1.25) {$n_{t_4}$};
\node (3) at (0,0) {$n_{t_3}$};

\node (5) at (-2,3) {$s_{l_1}$};
\node (6) at (-2,0.75) {$s_{l_2}$};

\node (7) at (2,1.875) {$\tilde{s}$};

\draw[->, line width=1pt] (5) to (1);
\draw[->, line width=1pt] (5) to (2);
\draw[->, dashed, line width=1pt] (5) to (3);
\draw[->, dashed, line width=1pt] (5) to (4);
\draw[->, dashed, line width=1pt] (6) to (2);
\draw[->, line width=1pt] (6) to (3);
\draw[->, line width=1pt] (6) to (4);

\draw[->, line width=1pt] (1) to (2);
\draw[->, dashed, bend left = 40, line width=1pt] (1) to (4);
\draw[->, line width=1pt] (2) to (4);
\draw[->, line width=1pt] (3) to (4);

\draw[->, line width=1pt] (1) to (7);
\draw[->, line width=1pt] (2) to (7);
\draw[->, line width=1pt] (3) to (7);
\draw[->, line width=1pt] (4) to (7);
\draw[->, bend left=40, in = 110, out = 80, looseness=1.5, line width=1pt] (5) to (7);
\draw[->, bend left=40, in = -110, out = -80, looseness=1.5, line width=1pt] (6) to (7);

\end{tikzpicture}

\end{minipage}

\caption{Example of the digraph construction.}
\label{Example_network}

\end{figure}

Let us present an example of this construction.
Consider two locations $l_1$ and $l_2$ with $travelTime^S(l_1,l_2) = 3$ and four tasks $t_1, t_2,t_3$ and $t_4$, whose attributes are displayed in the table on the left-hand side of Figure \ref{Example_network}.
The first row says that $t_1$ is a pickup task whose origin is $l_1$ at instant 1 and whose destination is also $l_1$ at instant 2.
Based on these tasks, we construct the digraph shown on the right-hand side of Figure \ref{Example_network}.
We indicate with a dashed-line arc when a shuttle trip is required between the tasks, and a solid-line arc otherwise.
In this digraph, for instance, there is not an arc between $n_{t_2}$ and $n_{t_3}$ since $t_2$ and $t_3$ are overlapped, and from $s_{l_2}$ to $n_{t_1}$ nor from $n_{t_1}$ to $n_{t_3}$ because the time gap between them is not enough to take the corresponding shuttle trip.
If, for example, a driver $d$ satisfies $l_d = l_1$ then $\langle t_1, t_2, t_4 \rangle$ is a driver route for $d$, which corresponds with the directed path $s_{l_1}, n_{t_{1}}, n_{t_{2}}, n_{t_{4}}, \tilde{s}$.
The cost of this path is 0 since no shuttles trips are required or, equivalently, no dashed-line arcs are traversed.

The structure of $G_{\delta}$ makes straightforward to prove the following remarks:

\begin{enumerate}
\item $G_{\delta}$ is a directed acyclic graph, i.e., a digraph with no directed cycles.
\item If $travelTime^S(p_t, q_t) \leq a_t$ for all $t \in T^t$, then the edge relation in $G_{\delta}$ is transitive, i.e., if $(u_1,u_2) \in \mathcal{A}$ and $(u_2,u_3) \in \mathcal{A}$ then $(u_1,u_3) \in \mathcal{A}$, for all $u_1,u_2,u_3 \in \mathcal{V}$. \label{TheoremEdgeTransitive}
\item There is a one-to-one correspondence between the driver routes for $d \in D$ and the $(s_{l_d},\tilde{s})$-directed paths in $G_{\delta}$. 
\end{enumerate}




The hypothesis of the second remark means that the shuttles are as fast as the trucks. As a consequence of the transitivity, removing a vertex from a directed path results in a new directed path.
The last remark allows us to define a feasible solution for the second stage of our problem as a pair, where the first component is a start time assignment $\delta$ for $T$ 
and the second one is a set with a $(s_{l_d},\tilde{s})$-directed path $P_d$ in $G_{\delta}$ for each $d \in D$ satisfying:
\begin{description}
    \item[Crew size constraints:] For all $t \in T$, $1 \leq |\{d \in D: n_t \in \mathcal{V}(P_d) \}| \leq 2$.
    \item[Rest constraints:] For all $d \in D$, $L_1$ is satisfied for $d$, i.e., $d$ rests at least 12 hours in every 24-hour interval
    and $d$ has at least one day off in every 7-day interval.
\end{description}

The objective value of a feasible solution $S = (\delta, \{P_d : d \in D\})$ is the total cost given by $w(S) = \sum_{d \in D} w(P_d)$,
and the aim of this stage is to find a solution $S$ that minimises $w(S)$; we refer to this value as $Z_3$.
In the appendix, we present an integer programming formulation that better describes our model, however a reader may omit it to understand
the rest of this work.

In the next section we present the main components of our algorithm, based on schemes of well-known metaheuristics, and we give details on how to implement them in terms of our specific problem. The reader may consult \cite{GENDREAU2019} for a more detailed explanation about general schemes of Greedy Randomised Adaptive Search Procedure (GRASP), Iterated Local Search (ILS), Variable Neighbourhood Descent (VND) and hybrid optimization approaches.

\section{Our algorithm}\label{OurAlgorithm}

Let us start by giving a brief description of how our algorithm works before moving on to more specific details. During the first stage, two methods can be applied: a direct one is to solve an integer linear programming formulation (see appendix)
which has the advantage that the given solution is optimal; the other is to provide a feasible solution via a heuristic which in our case
is semi-greedy. We give a brief explanation of this algorithm at the end of this section since this phase is beyond the scope of the current work.

For the second stage, we first construct a semi-greedy initial solution, whose drivers may work more time than allowed.
We then apply a repair procedure that, among other things, moves the start time of the tasks in conflict to minimise the workload of the drivers.
Once we reach a feasible solution, we focus on minimising the cost related to the external shuttle transports, using suitable moves to explore the space of feasible solutions.
After we reach a local optimum, we try to escape by applying a perturbation procedure that randomly changes the start time assignment without losing feasibility.
The best solution is kept and returned after a given time limit.

For the sake of clarity, we introduce our algorithm incrementally.
We first present a simple procedure that constructs a greedy feasible solution for our problem.
This procedure usually returns poor-quality solutions. 
Nevertheless, in subsequent steps, we incorporate new features that allow us to reach better results. These results will be later reported along with other computational experiments in Section \ref{SECTION-COMPUTATIONAL-EXPERIMENTS}.

\subsection{Constructing a greedy feasible solution}\label{Our-greedy-construction}

We follow a best insertion strategy to construct a greedy feasible solution: each task is inserted into the driver route that results in the lowest increase in the cost function; see the pseudo-code in Figure \ref{OurGreedyConstruction}.
At the beginning, the tasks are sorted by start time, and the driver route of each driver $d$ is represented as a directed path $P_d$ with the single node $s_{l_d}$.
In each iteration, an unplanned task $t$ is taken and a candidate list $CL$ is constructed with every driver $d$ such that $n_t$ can be inserted at the end of $P_d$, i.e., the augmentation of $P_{d}$ with $n_t$ is also a directed path and the rest constraints still hold for $d$.
After that, the driver $d^*$ of $CL$ that causes the smallest increase in the overall cost is chosen and $n_t$ is inserted at the end of $P_{d^*}$.
After all the tasks have been inserted, the sink node $\tilde{s}$ is inserted at the end of every directed path, leading to a feasible solution.
On the other hand, the algorithm \emph{fails} if some task remains unassigned. A repair procedure will be proposed later to deal with these cases.

It is clear that the greedy decisions we make during the construction may have led to a poor-quality solution.
For this reason, the first improvement is to use a search strategy to explore the space of feasible solutions to minimise the total cost associated with the external shuttle transports.

\begin{figure}
\centering
\begin{adjustbox}{minipage=[t]{.65\linewidth},fbox}
\begin{algorithmic}[1]
\Require Start time assignment $\delta$ for $T$
\Procedure{GREEDY}{}
\State sort $T$ by start time
\State $P_d \gets \{s_{l_d}\} \text{ for each } d \in D$ 
\ForAll {$t \in T$}
    \State $CL \gets \{d \in D: P_d \cup \{n_t\} \text{ is a directed path of } G_{\delta} ~ \wedge$ \hspace*{65pt} rest constraints hold for $d\}$
    \If {$CL = \emptyset$} fail \EndIf
    \State let $d^* \in CL$ such that $w(P_{d^*} \cup \{n_t\})$ is minimum
    \State $P_{d^*} \gets P_{d^*} \cup \{n_t\} $
\EndFor
\State $P_{d} \gets P_{d} \cup \{\tilde{s}\}$ for each $d \in D$
\State \Return $(\delta,\{P_d : d \in D\})$
\EndProcedure
\end{algorithmic}
\end{adjustbox}
\caption{Pseudo-code of our greedy construction.}
\label{OurGreedyConstruction}
\end{figure}

\subsection{Minimising the total shuttle cost} \label{Our-VND}

We use an iterative search to minimise the total shuttle cost of a feasible solution.
The moves that 
define the neighbourhood structures fall in two groups.
The first group of three moves has been extensively used in the literature, e.g., \cite{Shen2001,Ma2016}, and, in the context of our problem, they move one or more tasks from one driver route to another.
The second group of two moves is specific to our problem, and they change the size of a crew.
We explore each neighbourhood in order to find a lower-cost feasible solution, with respect to the cost function $Z_3$, and
we use a VND to deal with the multiple neighbourhood structures, e.g., \cite{HANSEN2001449,Hansen2010}.

Next, we give a comprehensive description of the aforementioned moves.
The input of each one is a feasible solution $S = (\delta, \{P_d: d \in D\})$ and they return a new solution $S' = (\delta, \{P'_d: d \in D\})$.
We remark that $S'$ might be infeasible (and in that case, the new solution is not considered).
The conditions that $S'$ must meet in order to be feasible are: (i) $P'_d$ must be a $(s_{l_d},\tilde{s})$-directed path of $G_{\delta}$ for all $d \in D$, (ii) crew size constrains must be satisfied for all tasks, and (iii) rest constraints must be satisfied for all drivers. The first two conditions are checked before the movement is performed, while (iii) is checked once $S'$
is constructed.
If $S'$ becomes infeasible or its cost is not better than $w(S)$, $S'$ is discarded.
In Figure \ref{FIGURE-MOVES}, we display a scheme of each move. Here, the straight lines with a cross in the middle represent those edges removed from a path during the move, and the dotted lines represent those edges added to a path after the move is performed.

\begin{description}

\item[\textsc{Move-1:} relocating a task.] For $d, d' \in D$, \emph{relocate} a task node $n_{t'}$ from $P_{d'}$ to $P_d$ between a couple of consecutive nodes $u$ and $v$ (see Figure \ref{FIGURE-MOVE-1}).

\medskip

\item[\textsc{Move-2}: swapping two task.] For $d, d' \in D$, \emph{swap} a task node $n_t$ from $P_d$ with a task node $n_{t'}$ from $P_{d'}$ (see Figure \ref{FIGURE-MOVE-2}).

\medskip

\item[\textsc{Move-3}: swapping two arcs.] For $d, d' \in D$, \emph{swap} the arc prior to a task node $n_t$ from $P_d$ with the arc prior to a task node $n_{t'}$ from $P_{d'}$, i.e., break the paths at those arcs and reconnect the first part of each one with the second part of the other (see Figure \ref{FIGURE-MOVE-3}).

\medskip

\item[\textsc{Move-4}: inserting a task.] For $d \in D$ and $t \in T$, \emph{insert} a task node $n_t$ into $P_d$ between a couple of consecutive nodes $u$ and $v$ (see Figure \ref{FIGURE-MOVE-4}).

\medskip

\item[\textsc{Move-5}: removing a task.] For $d \in D$, \emph{remove} a task node $n_t$ from $P_d$ (see Figure \ref{FIGURE-MOVE-5}).

\end{description}

\begin{figure}

\begin{minipage}{0.49\textwidth}
\begin{subfigure}{\textwidth}
\centering
  
\begin{tikzpicture}[every node/.style={draw, circle, minimum size=0.6cm, inner sep=0pt, line width=1pt}, scale = 0.75]

\node[draw = none] at (-5,0) {$P_d:$};
\node (A0) at (-4,0) {$s_{l_d}$};
\node (A1) at (-2,0) {$u$};
\node (A3) at (2,0) {$v$};
\node (A4) at (4,0) {$\tilde{s}$};

\def\x{-3}
\def\y{3}

\draw[->, line width=1pt] (A0) -- (\x-0.25,0);
\node[draw=none] at (\x,0) {\ldots};
\draw[->, line width=1pt]  (\x+0.25,0) -- (A1);
\draw[->, line width=1pt] (A1) -- (A3);
\node[draw=none] at (0,0) {\Large $\times$};
\draw[->, line width=1pt] (A3) -- (\y-0.25,0);
\node[draw=none] at (\y,0) {\ldots};
\draw[->, line width=1pt]  (\y+0.25,0) -- (A4);

\node[draw = none] at (-5,-1.5) {$P_{d'}:$};
\node (B0) at (-4,-1.5) {$s_{l_{d'}}$};
\node (B1) at (-2,-1.5) {$u'$};
\node (B2) at (0,-1.5) {$n_{t'}$};
\node (B3) at (2,-1.5) {$v'$};
\node (B4) at (4,-1.5) {$\tilde{s}$};

\def\x{-3}
\def\y{3}

\draw[->, line width=1pt] (B0) -- (\x-0.25,-1.5);
\node[draw=none] at (\x,-1.5) {\ldots};
\draw[->, line width=1pt]  (\x+0.25,-1.5) -- (B1);
\draw[->, line width=1pt] (B1) -- (B2);
\node[draw=none] at (-1,-1.5) {\Large $\times$};
\draw[->, line width=1pt] (B2) -- (B3);
\node[draw=none] at (1,-1.5) {\Large $\times$};
\draw[->, line width=1pt] (B3) -- (\y-0.25,-1.5);
\node[draw=none] at (\y,-1.5) {\ldots};
\draw[->, line width=1pt]  (\y+0.25,-1.5) -- (B4);

\draw[->, line width=1pt, densely dotted] (A1.south east) -- (B2.north west);
\draw[->, line width=1pt, densely dotted] (B2.north east) -- (A3.south west);

\draw[line width=1pt, densely dotted] (B1) -- (-2,-2.25);
\draw[line width=1pt, densely dotted] (-2,-2.25) -- (2,-2.25);
\draw[->, line width=1pt, densely dotted] (2,-2.25) -- (B3);

\end{tikzpicture}

\caption{Relocating a task.}
\label{FIGURE-MOVE-1}
\end{subfigure}

\medskip

\begin{subfigure}{\textwidth}
  \centering

\begin{tikzpicture}[every node/.style={draw, circle, minimum size=0.6cm, inner sep=0pt, line width=1pt}, scale = 0.75]

\node[draw = none] at (-5,0) {$P_d:$};
\node (A0) at (-4,0) {$s_{l_d}$};
\node (A1) at (-2,0) {$u$};
\node (A1) at (-2,0) {$u$};
\node (A2) at (0,0) {$n_t$};
\node (A3) at (2,0) {$v$};
\node (A4) at (4,0) {$\tilde{s}$};

\def\x{-3}
\def\y{3}

\draw[->, line width=1pt] (A0) -- (\x-0.25,0);
\node[draw=none] at (\x,0) {\ldots};
\draw[->, line width=1pt]  (\x+0.25,0) -- (A1);
\draw[->, line width=1pt] (A1) -- (A2);
\node[draw=none] at (-1,0) {\Large $\times$};
\draw[->, line width=1pt] (A2) -- (A3);
\node[draw=none] at (1,0) {\Large $\times$};
\draw[->, line width=1pt] (A3) -- (\y-0.25,0);
\node[draw=none] at (\y,0) {\ldots};
\draw[->, line width=1pt]  (\y+0.25,0) -- (A4);

\node[draw = none] at (-5,-1.5) {$P_{d'}:$};
\node (B0) at (-4,-1.5) {$s_{l_{d'}}$};
\node (B1) at (-2,-1.5) {$u'$};
\node (B2) at (0,-1.5) {$n_{t'}$};
\node (B3) at (2,-1.5) {$v'$};
\node (B4) at (4,-1.5) {$\tilde{s}$};

\def\x{-3}
\def\y{3}

\draw[->, line width=1pt] (B0) -- (\x-0.25,-1.5);
\node[draw=none] at (\x,-1.5) {\ldots};
\draw[->, line width=1pt]  (\x+0.25,-1.5) -- (B1);
\draw[->, line width=1pt] (B1) -- (B2);
\node[draw=none] at (-1,-1.5) {\Large $\times$};
\draw[->, line width=1pt] (B2) -- (B3);
\node[draw=none] at (1,-1.5) {\Large $\times$};
\draw[->, line width=1pt] (B3) -- (\y-0.25,-1.5);
\node[draw=none] at (\y,-1.5) {\ldots};
\draw[->, line width=1pt]  (\y+0.25,-1.5) -- (B4);

\draw[->, line width=1pt, densely dotted] (A1.south east) -- (B2.north west);
\draw[->, line width=1pt, densely dotted] (B2.north east) -- (A3.south west);
\draw[->, line width=1pt, densely dotted] (B1.north east) -- (A2.south west);
\draw[->, line width=1pt, densely dotted] (A2.south east) -- (B3.north west);

\end{tikzpicture}

  \caption{Swapping two tasks.}
  \label{FIGURE-MOVE-2}
\end{subfigure}

\end{minipage}
\begin{minipage}{0.49\textwidth}

\begin{subfigure}{\textwidth}
  \centering
\begin{tikzpicture}[every node/.style={draw, circle, minimum size=0.6cm, inner sep=0pt, line width=1pt}, scale = 0.75]

\node[draw = none] at (-5,0) {$P_d:$};
\node (A0) at (-4,0) {$s_{l_d}$};
\node (A1) at (-1.5,0) {$u$};
\node (A3) at (1.5,0) {$n_t$};
\node (A4) at (4,0) {$\tilde{s}$};

\def\x{-3}
\def\y{3}

\draw[->, line width=1pt] (A0) -- (\x-0.25,0);
\node[draw=none] at (\x,0) {\ldots};
\draw[->, line width=1pt]  (\x+0.25,0) -- (A1);
\draw[->, line width=1pt] (A1) -- (A3);
\node[draw=none] at (0,0) {\Large $\times$};
\draw[->, line width=1pt] (A3) -- (\y-0.25,0);
\node[draw=none] at (\y,0) {\ldots};
\draw[->, line width=1pt]  (\y+0.25,0) -- (A4);

\node[draw = none] at (-5,-1.5) {$P_{d'}:$};
\node (B0) at (-4,-1.5) {$s_{l_{d'}}$};
\node (B1) at (-1.5,-1.5) {$u'$};
\node (B3) at (1.5,-1.5) {$n_{t'}$};
\node (B4) at (4,-1.5) {$\tilde{s}$};

\def\x{-3}
\def\y{3}

\draw[->, line width=1pt] (B0) -- (\x-0.25,-1.5);
\node[draw=none] at (\x,-1.5) {\ldots};
\draw[->, line width=1pt]  (\x+0.25,-1.5) -- (B1);
\draw[->, line width=1pt] (B1) -- (B3);
\node[draw=none] at (0,-1.5) {\Large $\times$};
\draw[->, line width=1pt] (B3) -- (\y-0.25,-1.5);
\node[draw=none] at (\y,-1.5) {\ldots};
\draw[->, line width=1pt]  (\y+0.25,-1.5) -- (B4);

\draw[->, line width=1pt, densely dotted] (A1.south east) -- (B3.north west);
\draw[->, line width=1pt, densely dotted] (B1.north east) -- (A3.south west);

\end{tikzpicture}

  \caption{Swapping two arcs.}
  \label{FIGURE-MOVE-3}
\end{subfigure}

\medskip

\begin{subfigure}{\textwidth}
  \centering

\begin{tikzpicture}[every node/.style={draw, circle, minimum size=0.6cm, inner sep=0pt, line width=1pt}, scale = 0.75]

\node[draw = none] at (-5,0) {$P_d:$};
\node (A0) at (-4,0) {$s_{l_d}$};
\node (A1) at (-2,0) {$u$};
\node (A2) at (0,0.75) {$n_t$};
\node (A3) at (2,0) {$v$};
\node (A4) at (4,0) {$\tilde{s}$};

\def\x{-3}
\def\y{3}

\draw[->, line width=1pt] (A0) -- (\x-0.25,0);
\node[draw=none] at (\x,0) {\ldots};
\draw[->, line width=1pt]  (\x+0.25,0) -- (A1);
\draw[->, line width=1pt] (A1) -- (A3);
\node[draw=none] at (0,0) {\Large $\times$};
\draw[->, line width=1pt] (A3) -- (\y-0.25,0);
\node[draw=none] at (\y,0) {\ldots};
\draw[->, line width=1pt]  (\y+0.25,0) -- (A4);

\draw[line width=1pt, densely dotted] (A1) -- (-2,0.75);
\draw[->, line width=1pt, densely dotted] (-2,0.75) -- (A2);
\draw[line width=1pt, densely dotted] (A2) -- (2,0.75);
\draw[->, line width=1pt, densely dotted] (2,0.75) -- (A3);

\end{tikzpicture}

  \caption{Inserting a task.}
  \label{FIGURE-MOVE-4}
\end{subfigure}

\medskip

\begin{subfigure}{\textwidth}
  \centering

\begin{tikzpicture}[every node/.style={draw, circle, minimum size=0.6cm, inner sep=0pt, line width=1pt}, scale = 0.75]

\node[draw = none] at (-5,0) {$P_d:$};
\node (A0) at (-4,0) {$s_{l_d}$};
\node (A1) at (-2,0) {$u$};
\node (A2) at (0,0) {$n_t$};
\node (A3) at (2,0) {$v$};
\node (A4) at (4,0) {$\tilde{s}$};

\def\x{-3}
\def\y{3}

\draw[->, line width=1pt] (A0) -- (\x-0.25,0);
\node[draw=none] at (\x,0) {\ldots};
\draw[->, line width=1pt]  (\x+0.25,0) -- (A1);
\draw[->, line width=1pt] (A1) -- (A2);
\node[draw=none] at (-1,0) {\Large $\times$};
\draw[->, line width=1pt] (A2) -- (A3);
\node[draw=none] at (1,0) {\Large $\times$};
\draw[->, line width=1pt] (A3) -- (\y-0.25,0);
\node[draw=none] at (\y,0) {\ldots};
\draw[->, line width=1pt]  (\y+0.25,0) -- (A4);

\draw[line width=1pt, densely dotted] (A1) -- (-2,0.75);
\draw[line width=1pt, densely dotted] (-2,0.75) -- (2,0.75);
\draw[->, line width=1pt, densely dotted] (2,0.75) -- (A3);

\end{tikzpicture}

  \caption{Removing a task.}
  \label{FIGURE-MOVE-5}
\end{subfigure}
\end{minipage}

\caption{Moves.}
\label{FIGURE-MOVES}
\end{figure}

Up to this point, we have an algorithm that constructs a greedy feasible solution and moves iteratively to a lower-cost feasible solution following a VND strategy, until a local optimum (with respect to each of the neighbourhood structures) is reached.
We remark that this algorithm is deterministic, i.e., each execution converges to the same local optimum.
It is clear that the chances of guiding the search to a near-global optimum are increased when starting from different starting points.
For this reason, the feature presented in the next subsection is aimed to introduce non-determinism during the construction of the initial solution.

\subsection{Constructing semi-greedy feasible solutions}\label{Our-GRASP}

A semi-greedy construction can be easily derived from the greedy construction of Section \ref{Our-greedy-construction}: instead of inserting each task into the best driver route, we make a random decision among those that result in a cost contribution within a rate $\alpha$ of the best improvement.
To do so, we construct a restricted candidate list $RCL$ from $CL$ using a quality-based parameter $\alpha \in [0,1]$,
where $\alpha = 0$ stands for a behaviour exactly as the greedy construction of Section \ref{Our-greedy-construction}, and $\alpha = 1$
means a pure random construction (see \cite{Festa2009, Resende2016}), and we choose the driver $d^* \in RCL$ randomly.

Now, we define a GRASP based algorithm using the previous semi-greedy construction for the construction phase and the VND search of Section \ref{Our-VND} for the local search phase:
if the semi-greedy construction fails, i.e., there is a task with an empty candidate list, the partial solution obtained so far is dismissed
and the semi-greedy algorithm is started over.

Based on preliminary tests, we noted that the rest constraints are sometimes hard to meet and constructing feasible solutions becomes troublesome.
This happens more frequently in instances where the number of drivers is tight, since the drivers have greater workloads.
A well-known approach for this drawback is to allow infeasible solutions and minimise the infeasibility through the cost function; this approach is addressed by the next feature.

\subsection{Handling semi-feasible solutions}\label{SECTION-HANDLING-semi-FEASIBLE}

We address the problem of constructing feasible solutions with a two-stage procedure.
First, we allow to construct solutions where the drivers may violate one of the rest constraints given by $L_1$, specifically, now a driver may rest less than 12 hours in every 24-hour interval. We call it \emph{semi-feasible}. 
After that, we apply a repair procedure that tries to make it feasible.
The first stage is a modification of the semi-greedy construction, and the second one is an iterative search that minimises the infeasibility through an adequate cost function.
Before giving further details, let us introduce a new cost function to rank the semi-feasible solutions.

We refer to $\Gamma_{di}$ as the number of hours that a driver $d$ works in the 24-hours interval $[i,i+24]$ with $0 \leq i \leq 24H - 24$ (we consider that the time is divided into units of 1 hour, so $24H$ is the total number of hours of the planning horizon). 
Observe that $\Gamma_{di}$ can be easily computed from the driver route and the start time assignment (we recall that $\delta$ also sets the start time of the shuttle trips, since we assumed that they are scheduled as late as possible).
Now, given a semi-feasible solution $S = (\delta,\{P_d:d \in D\})$, we define the cost function:
\begin{equation}
    w_{inf}(S) =  \sum_{d \in D} \sum_{i=0}^{24H-24} \max \{0, \Gamma_{di}-12\} \label{infcost}
\end{equation}
For each $d \in D$, the term in the summation quantifies the total number of extra hours that $d$ works in each 24-hours interval.
It is easy to see that $w_{inf}(S) = 0$ if and only if $S$ is feasible.

Note that our definition of semi-feasible solution still does not allow a driver to work every day of the week.
Previous experimentation indicates that solutions generated by the semi-greedy violate this restriction infrequently.


We are now able to explain how the semi-feasible solutions are constructed and repaired.\\

\noindent \emph{Constructing semi-feasible solutions.} 
We modify the semi-greedy algorithm of Section \ref{Our-GRASP} to allow the construction of semi-feasible solutions.
The main idea is that we may insert a task into a driver route that introduces infeasibility (violations of the rest constraints) when no other options are available.
However, we use the cost function $w_{inf}$ to guide the construction towards good-quality semi-feasible solutions.
The pseudo-code of the new algorithm is shown in Figure \ref{PSEUDO-CODE-CONSTRUCTING-semi-FEASIBLE-SOLUTIONS}.
We introduce the following changes with respect to the original algorithm: the choice of $d^*$ depends on whether the candidate list is empty or not.
If so, we build a second candidate list $CL_{inf}$ with every driver $d$ such that $P_d \cup \{n_t\}$ is a directed path of $G_{\delta}$, without requiring that the 12-hour rest constraints hold.
Then, we choose $d^* \in CL_{inf}$ such that $w_{inf}(\delta, P_{d^*} \cup \{n_t\})$ is minimum.
If $CL \neq \emptyset$, we proceed as the original semi-greedy construction.

\begin{figure}
\centering
\begin{adjustbox}{minipage=[t]{.7\linewidth},fbox}
\begin{algorithmic}[1]
\Require Start time assignment $\delta$ for $T$, $\alpha \in [0,1]$
\Procedure{SEMI-GREEDY}{}
\State sort $T$ by start time
\State $P_d \gets \{s_{l_d}\} \text{ for each } d \in D$ 
\ForAll {$t \in T$}
    \State $CL \gets \{d \in D: P_d \cup \{n_t\} \text{ is a directed path of } G_{\delta} $ 
    \State \hskip 33pt $\wedge$ rest constraints hold for $d\}$
    \If {$CL = \emptyset$}
        \State $CL_{inf} \gets \{d \in D: P_d \cup \{n_t\} \text{ is a directed path of } G_{\delta}$
        \State \hskip 48pt $\wedge$ 1-day off constraints hold for $d\}$
        \State let $d^* \in CL_{inf}$ such that $w_{inf}(\delta, P_{d^*} \cup \{n_t\})$ is minimum 
    \Else
        \State make $RCL$ from $CL$ with quality-based parameter $\alpha$
        \State choose a random $d^* \in RCL$  
    \EndIf
    \State $P_{d^*} \gets P_{d^*} \cup \{n_t\} $
\EndFor
\State $P_{d} \gets P_{d} \cup \{\tilde{s}\}$ for each $d \in D$
\State \Return $(\delta,\{P_d : d \in D\})$
\EndProcedure
\end{algorithmic}
\end{adjustbox}
\caption{Pseudo-code of our semi-greedy construction to return semi-feasible solutions.}
\label{PSEUDO-CODE-CONSTRUCTING-semi-FEASIBLE-SOLUTIONS}
\end{figure}

The previous construction returns a semi-feasible solution $S = (\delta, \{P_d: d \in D\})$.
It is clear that if $w_{inf}(S) = 0$, then $S$ is already a feasible solution and nothing else should be done; otherwise, $S$ should be repaired.
A further description of how this repair procedure works is given below.\\

\noindent \emph{Repairing semi-feasible solutions.} 
The repair procedure uses an iterative search to minimise the cost function $w_{inf}$ given in (\ref{infcost}).
Again, the moves that define the neighbourhood structures fall in two groups. 
The first group has the five moves already presented in Section \ref{Our-VND}, which modify the driver routes and preserve the start time assignment.
The second group has a single move which, conversely, modifies the start time of a task and preserves the driver routes.
We explore the neighbourhoods defined by these moves to find a lower-cost semi-feasible solution, with respect to the cost function $w_{inf}$. We use a VND to deal with the multiple neighbourhood structures.

\begin{figure}
    \centering
\includegraphics[scale=0.50]{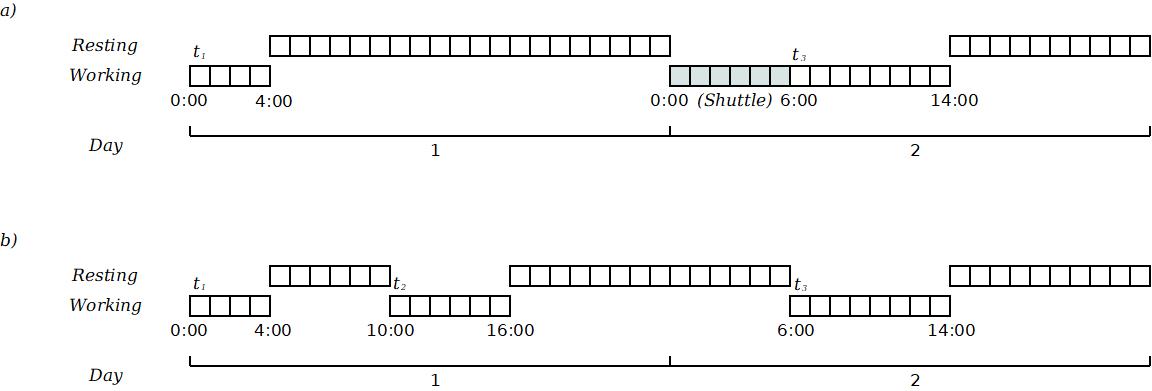}
    \caption{a) A semi-feasible solution,~ b) repairing the infeasibility.}
    \label{FIGHor2}
\end{figure}

It is easy to see that all these moves can help reducing infeasibility, even the insertion of a task.
Before proceeding, let us give an example of this. 
Suppose that there are three tasks: $t_1$ whose destination is $l_1$ and its duration is 4 hours, $t_2$ from $l_1$ to $l_2$ of 6 hours long, and $t_3$ whose origin is $l_2$ and its duration is 8 hours, such that $travelTime^S(l_1,l_2) = 6$, $\delta(t_1) = 0$, $\delta(t_2) = 10$ and $\delta(t_3) = 30$.
Additionally, suppose that a driver route has both $t_1$ and $t_3$, one after the other. 
It is clear that such driver works more than 12 hours in the 24-hour interval $[24,48]$, since the shuttle from $l_1$ to $l_2$ is scheduled immediately before $t_3$, see Figure \ref{FIGHor2}(a).
Now, if the task $t_2$ is inserted into this route between $t_1$ and $t_3$, then the shuttle is no longer needed and the previous infeasibility is repaired, as illustrated in Figure \ref{FIGHor2}(b).

Let us present the formal definition of the new move. The input is a semi-feasible solution $S = (\delta, \{P_d: d \in D\})$ and the output is another solution $S' = (\delta', \{P_d: d \in D\})$.
We remark that $S'$ might be infeasible (and in that case, it will not be accepted as the next current solution).
The conditions that $S'$ must meet in order to be semi-feasible are:
(i) $\delta'$ must be a start time assignment for $T$,
(ii) $P_d$ must be a $(s_{l_d},\tilde{s})$-directed path of $G_{\delta'}$ for all $d \in D$, and
(iii) 1-day off constraints must be satisfied for all drivers.

\begin{description}
    \item[\textsc{Move-6}: changing the start time of a task.]
    For $t \in T$ and a new start time $i$ for $t$, \emph{change} the value of $\delta(t)$ to $i$.
    
    As we have anticipated in Section \ref{SECTION-SOLUTION-APPROACH}, there is a dynamic time window with the tentative start times for $t$, delimited by the contiguous tasks in the truck route and the pickup/delivery time windows. 
    By choosing $i$ in this time window, we ensure that condition (i) is met.
    More explicitly, the value of $i$ must verify:
    \begin{itemize}
    
        \item Let $v \in V$ be the truck whose route contains $t$. If $t'$ is the task that precedes $t$ in $tr_v$, then $i \geq \delta(t') + a_{t'}$; if $t''$ is the task that follows $t$ in $tr_v$, then $i \leq \delta(t'') - a_t$. 
        And,
    
        \item if $t \in T^p \cup T^d$, then $i$ must belong to the respectively pickup/delivery time window of $t$.
        In addition, we do not allow to change the day in which the items are delivered to preserve the cost of the solution (recall that there is a penalty for the delay in the deliveries).
    
    \end{itemize}
    
    Besides, the task $t$ should be performed by the same crew (since this move does not modify the driver routes), so the value of $i$ is also delimited by the contiguous tasks in the driver routes.
    In order to meet (ii), the value of $i$ must also verify:

    \begin{itemize}

        \item Let $d \in D$ be a driver whose route contains $t$ (there are at most two of such drivers).
        If $t$ is the first task of $dr_v$, then $i \geq travelTime^S(l_d, p_t)$. If $t'$ is the task that precedes $t$ in $dr_v$, then $i \geq \delta(t') + a_{t'} + travelTime^S(q_{t'}, p_t)$; if $t''$ is the task that follows $t$ in $dr_v$, then $i \leq \delta(t'') - a_t - travelTime^S(q_t, p_{t''})$.
    \end{itemize}

\end{description}  

Again, the condition (iii) is checked once $S'$ is constructed.

As we have said before, in our GRASP based algorithm, we apply the repair procedure after the semi-greedy construction
provides a semi-feasible (but non feasible) solution.
When the repair procedure is unable to find a feasible solution, i.e., when it gets stuck in a local optimum $S'$ with
$w_{inf}(S') > 0$, we dismiss the current solution and continue with the next GRASP iteration.
Once we obtain a feasible solution, we run the VND search of Section \ref{Our-VND} to minimise the total shuttle cost.

Preliminary tests evidence that the repair procedure is often too time-consuming and, even worse, it does not always succeed.
Therefore, when a feasible solution is reached, it should be exploited as much as possible before constructing and repairing a new semi-feasible solution.
Next, we design a perturbation procedure that takes advantage of a current feasible solution to generate others.

\subsection{Perturbing feasible solutions}\label{SECTION-PERTUBING-SOLUTIONS}

Notice that applying \textsc{Move-6} to a feasible solution does not change its cost with respect to $Z_3$ (since the shuttle cost does not depend on the start time assignment), but new move opportunities might appear indeed.
A simple ILS, see e.g.~\cite{ILS,ILSApplications}, can be derived using this idea to escape from a local optimum.
Our perturbation procedure iterates over $t \in T$ in a random order and changes $\delta(t)$ for a random value among those that preserve feasibility (in case that no one preserves feasibility, $\delta(t)$ is not modified).

The full description of our algorithm results from embedding this perturbation procedure into the GRASP based algorithm.
The main idea is to perturb and then improve the solution a certain number of times before constructing a new semi-feasible solution; we give the pseudo-code in Figure \ref{PSEUDO-CODE-OUR-ALGORITHM}.

\begin{figure}
\centering
\begin{adjustbox}{minipage=[t]{.55\linewidth},fbox}
\begin{algorithmic}[1]
\Procedure{Our-GRASP}{}
\While {stopping criterion not satisfied}
\State $S \gets $ construct semi-feasible solution
\State repair $S$
\If {$S$ is feasible}
\State minimise the total shuttle cost of $S$
\Repeat
    \State perturb $S$
    \State minimise the total shuttle cost of $S$
    \State update best solution $S^*$
\Until{stopping criterion met}
\EndIf
\EndWhile
\State \Return $S^*$
\EndProcedure
\end{algorithmic}
\end{adjustbox}
\caption{Pseudo-code of our algorithm.}
\label{PSEUDO-CODE-OUR-ALGORITHM}
\end{figure}

\subsection{Heuristic resolution of first stage}

An inexpensive way to deliver truck routes is through a simple semi-greedy algorithm.
This algorithm iterates over the requests and inserts the corresponding pickup and delivery tasks into some truck route following a semi-greedy strategy. 
In order to prevent late deliveries and trucks from getting unavailable for a long time, the start time of the delivery task is always scheduled as early as possible and the pickup task as near to the delivery task as possible.
After this, the algorithm deals with the necessary trip tasks between the pickup and delivery tasks.
Given a pair of locations that needs a connection, the algorithm computes the shortest path between them and creates a trip task for each segment of the path (segments are delimited by the cities).
A random number is assigned to the start time of the trip tasks, as long as the truck route keeps synchronised.
We remark that the effectiveness of this simple algorithm comes from the fact that the truck routes are much easier to construct than the driver routes: trucks do not need to rest, and multiple-day time windows offer great flexibility.

\section{Computational experiments}\label{SECTION-COMPUTATIONAL-EXPERIMENTS}

In the present section we describe the computational experimentation carried out in order to test and evaluate the effectiveness of our approach.
For the sake of clarity, this section is divided as follows: in Section \ref{SECTION-INSTANCE-GENERATION} we explain the instances used to test the algorithms, 
in Section \ref{SECTION-CONFIG} we describe the configuration of the algorithms (e.g., values for the parameters, stopping criterion, and other implementation details), and in Section \ref{SECTION-RESULTS} we report and analyse the results.

\subsection{Instance generation}\label{SECTION-INSTANCE-GENERATION}

The problem we address comes from a coffee company in Colombia, and the constraints given in the present work are faithful to those of the original.
Unfortunately, it was not possible to access any information about the company's resources or instances.
For this reason, we decided to generate our own instances in a way consistent with reality and the nature of the problem.

\begin{figure}
    \centering
\includegraphics[scale=0.15]{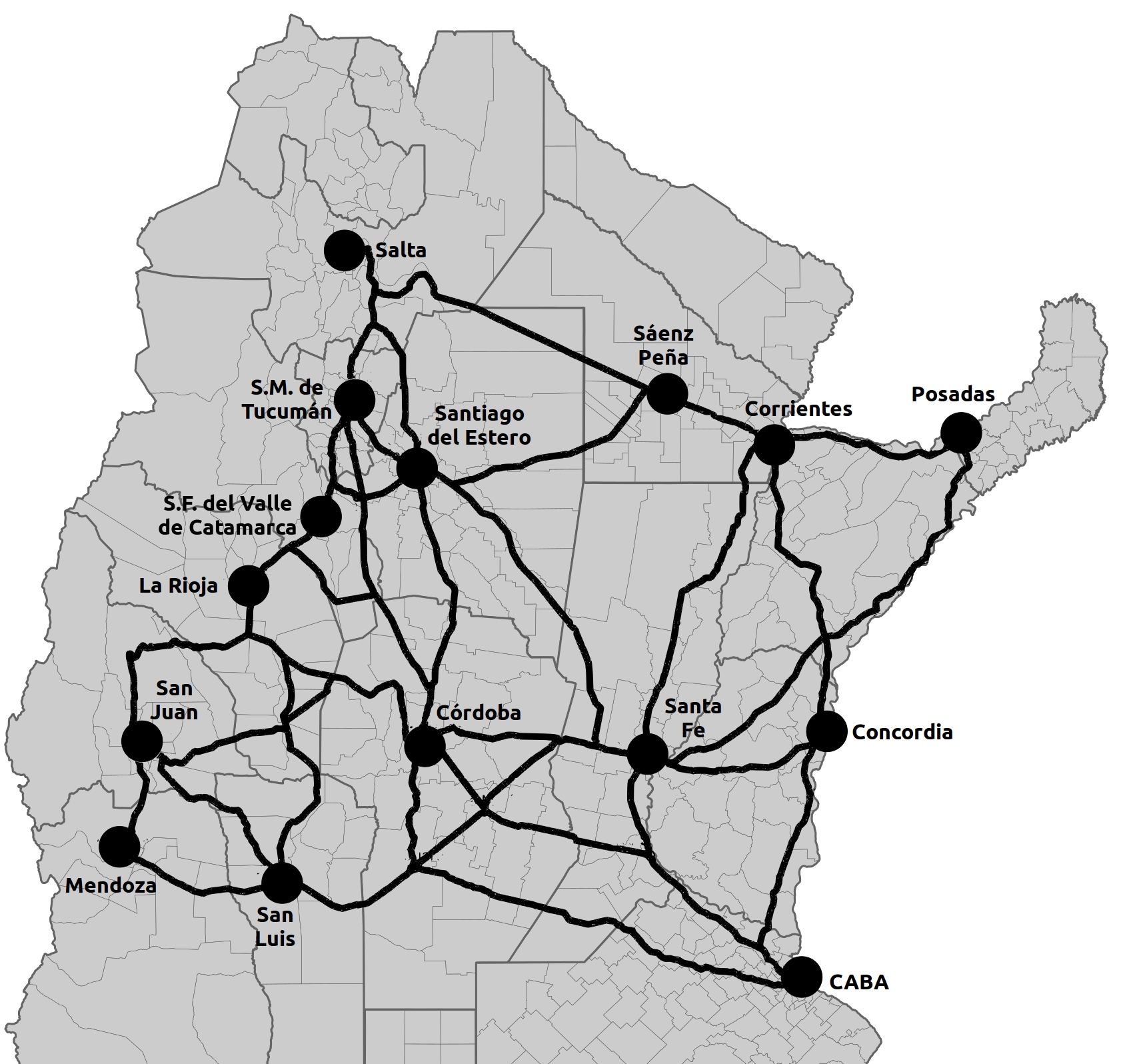}
    \caption{Cities and road network.}
    \label{FIGURE-MAP}
\end{figure}

We use as locations the 15 Argentine cities and the road network shown in Figure \ref{FIGURE-MAP}.
These cities meet several of the objectives sought: they are important urban agglomerations (indeed, almost all of them are capital cities, excepts Concordia and Sáenz Peña which are the second most important cities in their provinces), and there is a long distance between each pair of cities but not enough to force the drivers to rest en route (which is forbidden by the problem's constraints).
For a matter of road traffic safety, the road network only uses routes and highways of national jurisdiction, and the speed limit of the trucks and shuttles is set at 90km/h.
As a final remark, we are aware of the existence of local companies whose logistic is similar to the aforementioned one. 

We generate the instances with a routine that takes as incoming parameters:
the length of the planning horizon ($H$) and the number of requests (\#$R = |R|$),
trucks (\#$V = |V|$), and drivers (\#$D = |D|$).
Values are usually randomly generated integers with a uniform distribution in a given interval.
In particular, the attributes of each request are generated as follows:
\begin{itemize}
    \item Origin and destination: random but different cities.
    \item Pickup time window: the initial day $initDay_p$ is random in $[0,H-4]$, and the time interval has a random start $startTime_p$ in $[0,22]$ and a random ending in $[startTime_p,24]$.
    \item Delivery time window: the initial day is random in $[initDay_p,H-2]$, and the time interval has a random start $startTime_d$ in $[0,22]$ and a random ending in $[startTime_d,24]$.
\end{itemize}
Note that we do not let the initial days be at the end of the planning horizon since such requests would have few opportunities to be fulfilled, but a request whose initial day for delivery is set to $H-2$ can be still delivered at day $H$ (with the corresponding cost in $Z_1$).
The routine sets random initial locations for the trucks.
Then, it places a driver with a probability 0.8 in the initial location of each truck and sets random locations for the remaining drivers (assuming that there are more drivers than trucks).
Thus, there might be locations with trucks but without drivers (this happens, for example, when a truck-driver couple stops at some location and, subsequently, the driver is picked up by another truck), and there might be locations with drivers but without trucks (this happens, for example, when a truck with two drivers leaves one of them resting at some location).

\subsection{Algorithm configuration}\label{SECTION-CONFIG}

We have performed extensive preliminary experimentation to detect the most suitable configuration for each algorithm detailed in Section \ref{OurAlgorithm}.
This involves choosing values for the parameters, deciding the stopping criteria, and setting other implementation details such as the order of exploration for the iterative search.
For the sake of simplicity, we avoid reporting these preliminary tests and we directly present the chosen configurations, which are the ones that showed the best trade-off between quality of the output solutions and execution time.
We set $shuttleCost(l_1,l_2) = travelTime^S(l_1,l_2) + 1$ for every $l_1,l_2 \in L$.
Below, we enumerate the algorithms to be tested with their respective configurations and other implementation details.

\begin{description}

    \item[\textsc{Alg-1:}] GRASP presented in Section \ref{Our-GRASP}. 
    The construction phase uses $\alpha = 0.2$ to restrict the candidate lists.
    The VND uses a first-improving strategy, where the drivers are considered in non-increasing shuttle cost order, and the neighbourhoods are explored in the following order of moves: \textsc{Move-4}, \textsc{Move-5}, \textsc{Move-1}, \textsc{Move-2}, and \textsc{Move-3}.
    To speed-up the exploration, we first compute the cost of the neighbouring solution and, only when standing at a lower-cost one, we
    apply the concrete move and check feasibility (if the current neighbour turns out to be infeasible, the algorithm moves back to the last
    solution and the exploration proceeds).
    The descent stops when the cost cannot be further improved.

    \item[\textsc{Alg-2:}] \textsc{Alg-1} adapted to manage semi-feasible solutions (see Section \ref{SECTION-HANDLING-semi-FEASIBLE}).
    The construction phase uses $\alpha = 0.2$ when $d^*$ is chosen from $CL$ and a random tie-breaking rule when $d^*$ is chosen from $CL_{inf}$ (i.e., when the lowest increase for $w_{inf}$ occurs in more than one driver at the same time).
    The repair procedure uses a first-improving strategy, where the most infeasible drivers are considered first (i.e., those whose route has the highest value according to $w_{inf}$), and the neighbourhoods are explored in the following order of moves: \textsc{Move-4}, \textsc{Move-5}, \textsc{Move-6}, \textsc{Move-1}, \textsc{Move-2}, and \textsc{Move-3}.
    The repair procedure stops when there are no further cost improvements for $w_{inf}$.
    As this procedure is very time consuming, we only repair semi-feasible solutions $S$ whose cost $w_{inf}(S)$ does not exceed a threshold $\beta$, which is dynamically updated as follows.
    Initially, $\beta$ is set to $+\infty$ and is replaced with $w_{inf}(S_0)$ for the first semi-feasible solution $S_0$ that the algorithm achieves to repair. 
    Then, $\beta$ is increased by $1\%$ each time the construction phase gives a semi-feasible solution $S$ with $w_{inf}(S) > \beta$ (so the next one has greater chances to be repaired), and it is decreased by $1\%$ otherwise (so it gets more restrictive).

    \item[\textsc{Alg-3:}] \textsc{Alg-2} with the addition of the perturbation procedure (see Section \ref{SECTION-PERTUBING-SOLUTIONS}).
    The number of times that a feasible solution is perturbed is ruled by the formula $\lceil 25^{\gamma} \rceil$ with $\gamma = \frac{fails}{it+1}$, where $it$ is the number of the current GRASP iteration and $fails$ is the total number of GRASP iterations where no feasible solutions could be discovered.
    The resulting effect is that the number of perturbations grows exponentially when the algorithm systematically fails.
    Thus, the few feasible solutions found are exploited as much as possible to generate new ones.

\end{description}

\subsection{Results and analysis}\label{SECTION-RESULTS}

All the experiments are carried out in a desktop computer equipped with an Intel Core i5 2.6GHz processor (we use a single thread), 5.7GB of memory, Ubuntu 20.04 64bits operating system, GCC 5.4.0 compiler, and CPLEX 12.7 with default parameters (but using a single thread). 
All the implementations are written in the C++11 programming language.\\

\noindent \emph{First experiment}. 
We test \textsc{Alg-3} with the two approaches proposed for the first stage: the exact  resolution of the integer program and the semi-greedy heuristic.
We generate 3 instances of 100 requests with different planning horizons: 7, 14, and 28 days.
The number of trucks \#$V$ is respectively 32, 24, and 16.
This is because it takes more trucks to perform the same number of requests in a shorter planning horizon.
The number of drivers \#$D$ is 2(\#$V$) since, in order for a truck to be manned the whole day, two drivers are needed.
We introduce a parameter $\lambda \in [0,1]$ and we minimise the single-objective function
$$ Z_{12} = \lambda Z_1 + (1-\lambda) Z_2 $$
with 5 different values for $\lambda$: 0, 0.25, 0.5, 0.75, and 1.

We use CPLEX to solve the integer program for each value of $\lambda$, by fixing a time limit of 3600 seconds for each execution.
After that, we apply the CPLEX's \emph{populate} method to get 4 additional feasible solutions.
We configure this method to obtain solutions whose objective value is at most 3\% greater than the best solution found by CPLEX.
On the other hand, we run the semi-greedy heuristic 5 times for each value of $\lambda$. 
Thus, both methods return 5 different outputs for each instance and value of $\lambda$, and for each one we solve the second stage. 
The time limit for each run of $\textsc{Alg-3}$ is 900 seconds.

The results are reported in Table \ref{IP vs Semigreedy}.
The first 4 columns describe the instance parameters: number of days in the planning horizon, number of trucks, number of drivers, and the value of $\lambda$ in $Z_{12}$.
The next columns report the results obtained by the algorithms.
Column ``$Z_{12}^*$'' shows the value of $Z_{12}$ for the best feasible solution found by CPLEX,
``Time/Gap'' is the total elapsed time in \emph{seconds} during the optimization if the optimality is proven, or the final gap reported by CPLEX otherwise (the symbol ``\%'' is appended in that case),
``\#$T$'' is the average number of tasks,
``Sol.'' is the number of first stage outputs where $\textsc{Alg-3}$ finds some feasible solution.
The next 3 columns, ``$Z_3^{\text{min}}$'', ``$Z_3^{\text{avg}}$'' and ``$Z_3^{\text{max}}$'', display the minimum, average, and maximum value of $Z_3$, respectively, among the best feasible solutions found by $\textsc{Alg-3}$ for each first stage output,
``Fails'' is the average rate of fails (i.e. iterations where the construction phase fails to generate a feasible solution) of \textsc{Alg-3},
and ``Gap'' is the average gap between the costs of the solutions generated by the semi-greedy heuristic and the best solution found by CPLEX.
A symbol ``--'' means that no feasible solutions were discovered during the execution of \textsc{Alg-3}. When the values of ``$Z_3^{\text{min}}$'', ``$Z_3^{\text{avg}}$'', and ``$Z_3^{\text{max}}$''  are equal, we only report the one of ``$Z_3^{\text{avg}}$'' and leave a blank space in the others.

The table shows that the exact approach takes a considerable amount of time, and in particular it fails to prove optimality in 3 of the 15 executions within the time limit.
We also noted that this situation worsens when the instances grow in size (e.g. adding more requests), or when they grow in complexity (e.g. having a tighter number of trucks).
Moreover, high-quality solutions in the first stage do not necessarily lead to better solutions in the second stage; this fact is noticed in the instance with the longest planning horizon.
Conversely, the heuristic approach runs in a negligible amount of time and offers highly diversified solutions when runs more than once.
Because of these reasons, we will use the semi-greedy heuristic to deliver the truck routes to test our second-phase algorithm in the subsequent experiments.

When we emphasize $Z_1$ over $Z_2$ (i.e., $\lambda = 1$, only delays in deliveries are minimised), the second stage becomes harder to solve
(the percentage of fails is very high). We attribute it to the fact that there are more tasks and more overlap among them.
From now on, we fix $\lambda = 0.25$ which allows $Z_1$ and $Z_2$ to take part and provides a reasonable balance between $Z_3$ and percentage of fails in the second stage.

Note also that in some cases our algorithm is able to deliver optimal solutions for the second stage, precisely when $Z_3 = 0$, as no shuttles trip are needed.
This kind of solutions is of particular interest because they are also feasible on the \emph{no-shuttle} variant, i.e., where shuttle trips are not allowed at all.\\

\begin{table}[]
\small
\centering
\setlength{\tabcolsep}{2pt}
\begin{tabular}{
>{\centering}p{0.03\textwidth}
>{\centering}p{0.03\textwidth}
>{\centering}p{0.03\textwidth}
>{\centering}p{0.04\textwidth}|
>{\centering}p{0.06\textwidth}
>{\centering}p{0.06\textwidth}
>{\centering}p{0.04\textwidth}
>{\centering}p{0.03\textwidth}
>{\centering}p{0.04\textwidth}
>{\centering}p{0.04\textwidth}
>{\centering}p{0.04\textwidth}
>{\centering}p{0.04\textwidth}|
>{\centering}p{0.04\textwidth}
>{\centering}p{0.04\textwidth}
>{\centering}p{0.03\textwidth}
>{\centering}p{0.04\textwidth}
>{\centering}p{0.04\textwidth}
>{\centering}p{0.04\textwidth}
>{\centering\arraybackslash}p{0.04\textwidth}}
\toprule
\multirow{2}{*}{$H$} & \multirow{2}{*}{\#$V$} & \multirow{2}{*}{\#$D$} & \multirow{2}{*}{$\lambda$} & \multicolumn{8}{c|}{Exact + \textsc{Alg-3}} & \multicolumn{7}{c}{Heuristic + \textsc{Alg-3}}\\
&&&& $Z_{12}^*$ & Time/Gap & \#$T$ & Sol. & $Z_3^{\text{min}}$ & $Z_3^{\text{avg}}$ & $Z_3^{\text{max}}$ & Fails & Gap & \#$T$ & Sol. & $Z_3^{\text{min}}$ & $Z_3^{\text{avg}}$ & $Z_3^{\text{max}}$ & Fails\\ 
\midrule
\multirow{5}{*}{7} & \multirow{5}{*}{32} & \multirow{5}{*}{64} & 0 & 986 & 4 s & 437 & 5 & & 0 & & 24\% & 15\% & 490 & 3 & & 0 & & 54\%\\ 
  &  &  & 0.25 & 855.25 & 0.36\% & 443 & 5 & & 0 & & 46\% & 19\% & 491 & 3 & 0 & 2 & 6 & 63\%\\ 
  &  &  & 0.5 & 675 & 137 s & 452 & 5 & 0 & 2.4 & 6 & 39\% & 25\% & 538 & 1 & & 6 & & 96\%\\ 
  &  &  & 0.75 & 480.5 & 48 s & 454 & 3 & & 12 & & 77\% & 27\% & 591 & 0 & & -- & & 100\%\\ 
  &  &  & 1 & 283 & 4 s & 583 & 1 & & 12 & & 99\% & 22\% & 604 & 1 & & 16 & & 95\%\\
\midrule
\multirow{5}{*}{14} & \multirow{5}{*}{24} & \multirow{5}{*}{48} & 0 & 947 & 3 s & 445 & 5 & & 0 & & 55\% & 21\% & 512 & 4 & & 0 & & 37\%\\ 
  &  &  & 0.25 & 770 & 119 s & 452 & 5 & & 0 & & 38\% & 19\% & 508 & 4 & & 0 & & 59\%\\ 
  &  &  & 0.5 & 543 & 86 s & 458 & 5 & & 0 & & 82\% & 25\% & 534 & 5 & & 0 & & 63\%\\ 
  &  &  & 0.75 & 313 & 21 s & 457 & 5 & & 0 & & 89\% & 32\% & 589 & 4 & 0 & 3.8 & 15 & 92\%\\ 
  &  &  & 1 & 83 & 209 s & 622 & 3 & 4 & 6.3 & 9 & 99\% & 20\% & 631 & 4 & 8 & 18 & 25 & 99\%\\
\midrule
\multirow{5}{*}{28} & \multirow{5}{*}{16} & \multirow{5}{*}{32} & 0 & 1096 & 42 s & 456 & 5 & 30 & 41.2 & 65 & 86\% & 19\% & 518 & 5 & 16 & 25.4 & 42 & 72\%\\ 
  &  &  & 0.25 & 916.5 & 2.84\% & 473 & 5 & 9 & 14.6 & 20 & 70\% & 15\% & 521 & 5 & 6 & 18.4 & 33 & 73\%\\ 
  &  &  & 0.5 & 649.5 & 1.52\% & 478 & 5 & 6 & 15.4 & 27 & 67\% & 18\% & 547 & 5 & 9 & 23 & 34 & 76\%\\ 
  &  &  & 0.75 & 370 & 66 s & 480 & 5 & 6 & 11 & 16 & 71\% & 26\% & 594 & 4 & 25 & 37.8 & 47 & 91\%\\ 
  &  &  & 1 & 89 & 448 s & 619 & 5 & 27 & 51.2 & 70 & 97\% & 30\% & 624 & 5 & 24 & 31.2 & 41 & 87\%\\
\bottomrule
\end{tabular}
\caption{Comparison between the exact and the heuristic approaches to solve the first stage.}
\label{IP vs Semigreedy}
\end{table}

\noindent \emph{Second experiment}.
We compare the performance of \textsc{Alg-1}, \textsc{Alg-2}, and \textsc{Alg-3}.
We keep the instances of the first experiment and we consider more number of trucks, now
\#$V \in \{18,25,32\}$ for $H=7$,
\#$V \in \{13,18,24\}$ for $H=14$,
and \#$V \in \{8,12,16\}$ for $H=28$.
These values have emerged from: the least number of trucks such that the heuristic approach for the first stage is able to find a feasible solution, an intermediate value, and the number of trucks used in the first experiment.
We keep \#$D = 2$(\#$V$).

For each instance, we generate 3 different outputs from the first stage by running the semi-greedy heuristic.
Then, we solve the second stage with the mentioned algorithms, using 3600 seconds as time limit in each run (summarising a maximum of 81 hours of CPU time). These specifications (number of outputs and time limit) are fixed for the next experiments.

The results can be found in Table \ref{Alg-1 vs Alg-2 vs Alg-3}.
The column $Z_{12}^{\text{avg}}$ shows the average cost of the solutions generated by the semi-greedy heuristic,
``Iter'' is the average number of iterations in \emph{thousands} done in the second stage,
and ``BestT'' is the average time in \emph{seconds} needed to find the best solution in the second stage.

As it can be seen from the table, \textsc{Alg-1} finds at least one feasible solution in 5 of the 9 instances, but failing more that 90\% of the times in 3 of them.
This fact highlights the difficulty of generating feasible solutions for our problem.
\textsc{Alg-2} finds feasible solutions in more instances, precisely 7 of 9.
In particular, the repair procedure helps to reduce the rate of fails.
It is clear that \textsc{Alg-3} outperforms the others as it finds the highest-quality solutions.
Moreover, when \textsc{Alg-2} and \textsc{Alg-3} deliver solutions with the same cost, the latter finds it faster as evidenced by parameter BestT.
In summary, despite the additional features make the iterations more time-consuming, the whole algorithm becomes more effective.

Also note that the costs of the solutions (both $Z_{12}$ and $Z_3$) and the number of tasks decrease when more resources (trucks and drivers) are available.

Finally, we repeated this experiment with a longer time limit (10 hours) for the instances where no feasible solutions are discovered ($H=14$ and
\#$V=13$, and $H=28$ and \#$V=8$), but we obtained the same negative result.
We suspect that these instances are infeasible.\\

\begin{table}[]
\small
\centering
\setlength{\tabcolsep}{2pt}
\begin{tabular}{
>{\centering}p{0.02\textwidth}
>{\centering}p{0.02\textwidth}
>{\centering}p{0.03\textwidth}|
>{\centering}p{0.04\textwidth}
>{\centering}p{0.03\textwidth}|
>{\centering}p{0.025\textwidth}
>{\centering}p{0.035\textwidth}
>{\centering}p{0.035\textwidth}
>{\centering}p{0.03\textwidth}
>{\centering}p{0.04\textwidth}
>{\centering}p{0.04\textwidth}|
>{\centering}p{0.025\textwidth}
>{\centering}p{0.035\textwidth}
>{\centering}p{0.035\textwidth}
>{\centering}p{0.03\textwidth}
>{\centering}p{0.04\textwidth}
>{\centering}p{0.04\textwidth}|
>{\centering}p{0.025\textwidth}
>{\centering}p{0.035\textwidth}
>{\centering}p{0.035\textwidth}
>{\centering}p{0.03\textwidth}
>{\centering}p{0.04\textwidth}
>{\centering\arraybackslash}p{0.04\textwidth}}
\toprule
\multirow{2}{*}{$H$} & \multirow{2}{*}{\#$V$} & \multirow{2}{*}{\#$D$} & \multirow{2}{*}{$Z_{12}^{\text{avg}}$} & \multirow{2}{*}{\#$T$} & \multicolumn{6}{c|}{\textsc{Alg-1}} & \multicolumn{6}{c|}{\textsc{Alg-2}} & \multicolumn{6}{c}{\textsc{Alg-3}}\\
&&&&& Sol. & $Z_3^{\text{min}}$ & $Z_3^{\text{avg}}$ & Iter & BestT & Fails & Sol. & $Z_3^{\text{min}}$ & $Z_3^{\text{avg}}$ & Iter & BestT & Fails & Sol. & $Z_3^{\text{min}}$ & $Z_3^{\text{avg}}$ & Iter & BestT & Fails \\ 
\midrule
\multirow{3}{*}{7} & 18 & 36 & 1209 & 507 & 0 & -- & -- & 399 & -- & 100\% & 2 & 46 & 59 & 6 & 2887 & 99\% & 2 & 23 & 31.5 & 6 & 1239 & 99\%\\
 & 25 & 50 & 1078 & 494 & 1 & 7 & 7 & 293 & 490 & 99\% & 3 & 4 & 6.3 & 28 & 1082 & 79\% & 3 & 0 & 4.3 & 14 & 959 & 78\%\\
 & 32 & 64 & 1061 & 494 & 2 & 0 & 6 & 612 & 8 & 82\% & 2 & 0 & 6 & 18 & 18 & 50\% & 2 & 0 & 3 & 14 &  7 & 50\%\\
\midrule
\multirow{3}{*}{14} & 13 & 26 & 1144 & 538 & 0 & -- & -- & 494 & -- & 100\% & 0 & -- & -- & 1 & -- &100\% & 0 & -- & -- & 1 & -- & 100\%\\
 & 18 & 36 & 1027 & 520 & 2 & 10 & 27.5 & 214 & 2405 &99\% & 3 & 3 & 7 & 21 & 995 &78\% & 3 & 0 & 1.3 & 7 & 2025 & 78\%\\
 & 24 & 48 & 959 & 507 & 2 & 0 & 0 & 250 & 18 & 78\% & 2 & 0 & 0 & 6 & 137 & 67\% & 2 & 0 & 0 & 5 & 17 & 59\%\\
\midrule
\multirow{3}{*}{28} & 8 & 16 & 1257 & 556 & 0 & -- & -- & 1027 & -- & 100\% & 0 & -- & -- & 1 & -- & 100\% & 0 & -- & -- & 1 & -- & 100\%\\
 & 12 & 24 & 1139 & 532 & 0 & -- & -- & 635 & -- &100\% & 2 & 84 & 92.5 & 5 & 2524 &99\% & 2 & 59 & 70.5 & 5 & 1727 &99\%\\
 & 16 & 32 & 1086 & 521 & 3 & 22 & 29 & 120 & 832 &91\% & 3 & 24 & 25.7 & 18 & 2800 &66\% & 3 & 12 & 14.3 & 7 & 1032 &66\%\\
\bottomrule
\end{tabular}
\caption{Comparison between \textsc{Alg-1}, \textsc{Alg-2}, and \textsc{Alg-3}.}
\label{Alg-1 vs Alg-2 vs Alg-3}
\end{table}

\noindent \emph{Third experiment}. We analyse the impact of changing the number of drivers on the performance of \textsc{Alg-3}.
We select the most challenging instances of the previous experiment, i.e.~those with the lowest number of trucks for each planning horizon, and we consider the following number of drivers:
\#$D \in \{29,36,43,50\}$ for \#$V=18$,
\#$D \in \{26,31,36\}$ for \#$V=13$,       
and \#$D \in \{16,20,24\}$ for \#$V=8$.

The results are reported in Table \ref{Variando D}.
Now, feasible solutions can be found for the instances with \#$V=13$ and \#$V=8$
if \#$D$ is increased by 20\% and 50\% respectively,
but none for the instance with \#$V=18$ 
if \#$D$ is decreased by 20\%.
Also, the cost of the best solutions and the rate of fails diminish when there are more drivers.
We conclude that our algorithm performs better when more drivers are available.\\

    
\begin{table}[]
\begin{minipage}{0.4\linewidth}
    \small
    \centering
    \setlength{\tabcolsep}{2pt}
    \begin{tabular}{
    >{\centering}p{0.065\textwidth}
    >{\centering}p{0.065\textwidth}
    >{\centering}p{0.065\textwidth}|
    >{\centering}p{0.065\textwidth}
    >{\centering}p{0.085\textwidth}
    >{\centering}p{0.085\textwidth}
    >{\centering\arraybackslash}p{0.085\textwidth}}
    \toprule
    \multirow{2}{*}{$H$} & \multirow{2}{*}{\#$V$} & \multirow{2}{*}{\#$D$} & \multicolumn{4}{c}{\textsc{Alg-3}}\\
    & & & Sol. & $Z_3^{\text{min}}$ & $Z_3^{\text{avg}}$ & Fails\\ 
    \midrule
    \multirow{4}{*}{7} & \multirow{4}{*}{18} & 29 & 0 & -- & -- & 100\%\\
     & & 36 & 2 & 23 & 31.5 & 99\%\\
     & & 43 & 3 & 6 & 8.7 & 79\%\\
     & & 50 & 3 & 0 & 5.3 & 68\%\\
    \midrule
    \multirow{3}{*}{14} & \multirow{3}{*}{13} & 26 & 0 & -- & -- & 100\%\\
    & & 31 & 3 & 6 & 21 & 96\%\\
    & & 36 & 3 & 0 & 0 & 63\%\\
    \midrule
    \multirow{3}{*}{28} & \multirow{3}{*}{8} & 16 & 0 & -- & -- & 100\%\\
    & & 20 & 0 & -- & -- & 100\%\\
    & & 24 & 3 & 24 & 35.7 & 96\%\\
    \bottomrule
    \end{tabular}
    \caption{Changing the number of drivers.}
    \label{Variando D}
    \end{minipage}
    \hfill
\begin{minipage}{0.55\textwidth}
\small
\centering
\setlength{\tabcolsep}{2pt}
\begin{tabular}{
>{\centering}p{0.05\textwidth}
>{\centering}p{0.05\textwidth}
>{\centering}p{0.05\textwidth}|
>{\centering}p{0.05\textwidth}
>{\centering}p{0.07\textwidth}
>{\centering}p{0.07\textwidth}
>{\centering}p{0.07\textwidth}|
>{\centering}p{0.05\textwidth}
>{\centering}p{0.07\textwidth}
>{\centering}p{0.07\textwidth}
>{\centering\arraybackslash}p{0.07\textwidth}}
\toprule
\multirow{2}{*}{$H$} & \multirow{2}{*}{\#$V$} & \multirow{2}{*}{\#$D$} & \multicolumn{4}{c|}{1 driver} & \multicolumn{4}{c}{$\leq 2$ drivers}\\
&&& Sol. & $Z_3^{\text{min}}$ & $Z_3^{\text{avg}}$ & Fails & Sol. & $Z_3^{\text{min}}$ & $Z_3^{\text{avg}}$ & Fails\\ 
\midrule
\multirow{3}{*}{7} & 18 & 36 & 2 & 66 & 84 & 99\% & 2 & 23 & 31.5 & 99\%\\
 & 25 & 50 & 3 & 22 & 23.3 & 77\% & 3 & 0 & 4.3 & 78\%\\
 & 32 & 64 & 2 & 6 & 11 & 67\% & 2 & 0 & 3 & 50\%\\
\midrule
\multirow{3}{*}{14} & 13 & 31 & 3 & 61 & 84 & 96\% & 3 & 6 & 21 & 96\%\\
 & 18 & 36 & 3 & 18 & 21.3 & 77\% & 3 & 0 & 1.3 & 78\%\\
 & 24 & 48 & 2 & 5 & 9.5 & 67\% & 2 & 0 & 0 & 59\%\\
\midrule
\multirow{3}{*}{28} & 8 & 24 & 3 & 30 & 53.3 & 96\% & 3 & 24 & 35.7 & 96\%\\
 & 12 & 24 & 3 & 64 & 86.7 & 99\% & 2 & 59 & 70.5 & 99\%\\
 & 16 & 32 & 3 & 41 & 45 & 65\% & 3 & 12 & 14.3 & 66\%\\
 \bottomrule
\end{tabular}
\caption{Changing the crew size.}
\label{1 vs 2 drivers}
\end{minipage}
    
\end{table}

\noindent \emph{Fourth experiment}.
We test the performance of \textsc{Alg-3} when the crew size is restricted to a single driver.
This is achieved by disabling \textsc{Move-4} and \textsc{Move-5}, i.e., the moves that change the crew size.
We use the same instances as in the second experiment, but taking more drivers in the two instances where \textsc{Alg-3} does not find any feasible solution.
More precisely, as in the third experiment, we take \#$D=31$ for the instance with \#$V=13$ and \#$D=24$ for the one with \#$V=8$.

Table \ref{1 vs 2 drivers} presents the results.
Columns 4-7 use crews with size equal to 1, and columns 8-11, with size lower or equal than 2.
When \textsc{Move-4} and \textsc{Move-5} are active, there is a major improvement in the cost of the best solution for all the instances,
more precisely $Z_3^{\text{min}}$ is 60\% lower on average.
It is evident that carrying an additional driver is highly effective in reducing the dependence on shuttles.\\

\noindent \emph{Fifth experiment}. We test $\textsc{Alg-3}$ with additional rest regulations, where a driver cannot work more than 60 hours per week, and must rest uninterruptedly for at least 11 hours between two consecutive work periods.
We will refer to these variants as: $L_1+L_2$ and $L_1+L_3$, respectively.
Any semi-feasible solution that violates $L_2$ or $L_3$ is considered infeasible for these variants.
We select from the previous experiment the instances with the lowest \#$V$ for each planning horizon, and for each one we generate a
copy with $30\%$ more drivers (since stronger rest constraints are expected to require more drivers).
We also consider larger instances, of 1000 and 3000 requests, with corresponding reasonable amounts of trucks and drivers (we keep the same proportion of trucks per requests and \#$D \in \{1.9$\#$V$, 2.2\#$V\}$ to be precise).
Table \ref{L1 L2 L3} summarises the results. Columns 5-9 consider $L_1$, columns 10-14, $L_1 + L_2$, and columns 15-19, $L_1 + L_3$.
When ``Iter'' is lower than 0.05 (500 iterations), we write $>0.1$.

In the case of 100 requests, the performances are fairly similar when $L_2$ is active, in fact the algorithm finds feasible solutions of equal
cost for the instances with more drivers.
Concerning $L_3$, no feasible solutions are found unless there are more drivers.
In that case, $Z_3^{\text{min}}$ is as good as before for $H=14$ and worse for $H \in \{7,28\}$.
Also note that with $L_3$, the algorithm does many more iterations and fails more often; this exposes the difficulty of constructing feasible solutions
satisfying $L_3$.
Despite these exceptions, our algorithm seems to adapt to the different regulations.

For large instances and cases $L_1$ and $L_1 + L_2$, less drivers are needed per truck and, even so, they are fairly easy to solve,
e.g., an optimal solution is found in few iterations, also useful for the \emph{no-shuttle} variant.
In the case of $L_1 + L_3$, the same pattern (as for 100 requests) arises: more drivers are needed to find a feasible solution.

\begin{table}[]
\small
\centering
\setlength{\tabcolsep}{2pt}
\begin{tabular}{
>{\centering}p{0.05\textwidth}
>{\centering}p{0.03\textwidth}
>{\centering}p{0.03\textwidth}
>{\centering}p{0.04\textwidth}|
>{\centering}p{0.03\textwidth}
>{\centering}p{0.04\textwidth}
>{\centering}p{0.04\textwidth}
>{\centering}p{0.05\textwidth}
>{\centering}p{0.04\textwidth}|
>{\centering}p{0.03\textwidth}
>{\centering}p{0.04\textwidth}
>{\centering}p{0.04\textwidth}
>{\centering}p{0.05\textwidth}
>{\centering}p{0.04\textwidth}|
>{\centering}p{0.03\textwidth}
>{\centering}p{0.04\textwidth}
>{\centering}p{0.04\textwidth}
>{\centering}p{0.05\textwidth}
>{\centering\arraybackslash}p{0.04\textwidth}}
\toprule
\multirow{2}{*}{\#$R$} & \multirow{2}{*}{$H$} & \multirow{2}{*}{\#$V$} & \multirow{2}{*}{\#$D$} & \multicolumn{5}{c|}{$L_1$} & \multicolumn{5}{c|}{$L_1 + L_2$} & \multicolumn{5}{c}{$L_1 + L_3$}\\
&&&& Sol. & $Z_3^{\text{min}}$ & $Z_3^{\text{avg}}$ & Iter & Fails & Sol. & $Z_3^{\text{min}}$ & $Z_3^{\text{avg}}$ & Iter & Fails & Sol. & $Z_3^{\text{min}}$ & $Z_3^{\text{avg}}$ & Iter & Fails\\
\midrule
& \multirow{2}{*}{7} & \multirow{2}{*}{18} & 36 & 2 & 23 & 31.5 & 5 & 99\% & 1 & 30 & 30 & 5 & 99\% & 0 & -- & -- & 300 & 100\% \\
&  &  & 47 & 3 & 0 & 3.3 & 11 & 64\% & 3 & 0 & 3.3 & 11 & 63\% & 2 & 16 & 16.5 & 122 & 99\% \\
\cmidrule{2-19}
\multirow{2}{*}{100} & \multirow{2}{*}{14} & \multirow{2}{*}{13} & 31 & 3 & 6 & 21 & 7 & 96\% & 3 & 7 & 18.7 & 7 & 96\% & 0 & -- & -- & 109 & 100\% \\
&  &  & 40 & 3 & 0 & 0 & 0.1 & 67\% & 3 & 0 & 0 & 0.1 & 67\% & 3 & 0 & 9.3 & 20 & 97\% \\
\cmidrule{2-19}
& \multirow{2}{*}{28} & \multirow{2}{*}{8} & 24 & 3 & 24 & 35.7 & 6 & 96\% & 3 & 26 & 36.3 & 6 & 96\% & 0 & -- & -- & 149 & 100\% \\
&  &  & 31 & 3 & 12 & 15.7 & 6 & 81\% & 3 & 12 & 18.3 & 6 & 81\% & 3 & 37 & 55 & 53 & 99\% \\
\midrule
\multirow{4}{*}{1000} & \multirow{2}{*}{7} & \multirow{2}{*}{180} & 342 & 3 & 0 & 0 & 0.1 & 77\% & 3 & 0 & 0 & 0.2 & 77\% & 0 & -- & -- & 0.7 & 100\%\\
 &  &  & 396 & 3 & 0 & 0 & > 0.1 & 28\% & 3 & 0 & 0 & > 0.1 & 28\% & 3 & 0 & 0 & 0.1 & 72\%\\
\cmidrule{2-19}
 & \multirow{2}{*}{28} & \multirow{2}{*}{80} & 152 & 3 & 0 & 4.7 & 0.4 & 87\% & 3 & 0 & 4 & 0.3 & 91\% & 0 & -- & -- & 1.9 & 100\%\\
 &  &  & 176 & 3 & 0 & 0 & > 0.1 & 27\% & 3 & 0 & 0 & > 0.1 & 56\% & 2 & 3 & 6 & 1 & 98\%\\
\midrule
\multirow{4}{*}{3000} & \multirow{2}{*}{7} & \multirow{2}{*}{540} & 1026 & 3 & 0 & 0 & > 0.1 & 86\% & 3 & 0 & 0 & > 0.1 & 86\% & 0 & -- & -- & 0.1 & 100\%\\
 &  &  & 1188 & 3 & 0 & 0 & > 0.1 & 0\% & 3 & 0 & 0 & > 0.1 & 0\% & 2 & 0 & 0 & > 0.1 & 81\%\\
\cmidrule{2-19}
 & \multirow{2}{*}{28} & \multirow{2}{*}{240} & 456 & 3 & 0 & 0 & > 0.1 & 56\% & 3 & 0 & 0 & > 0.1 & 47\% & 2 & 20 & 20.5 & 0.1 & 99\%\\
 &  &  & 528 & 3 & 0 & 0 & > 0.1 & 25\% & 3 & 0 & 0 & > 0.1 & 20\% & 3 & 0 & 2 & > 0.1 & 58\%\\
\bottomrule
\end{tabular}
\caption{Using additional rest constraints and larger instances.}
\label{L1 L2 L3}
\end{table}

\section{Conclusions} \label{SECTION-CONCLU}

In the present work, we address a problem concerning the simultaneous routing and scheduling of a fleet of trucks and a staff of drivers to fulfill pickup-and-delivery requests over a planning horizon.
This problem has the particularity that the trucks play a double role: besides carrying the requests, they can be used to transport the drivers among the locations.

As it applies to long-distance road transportation, travel times usually exceed the working hours of the drivers,
so it becomes essential to allow relays during the trips
(as opposed to short-distance transportation where a vehicle can be manned by a single driver, who can make multiple requests within her/his working day).
On the other hand, considering more than one driver per crew allows a better use of the fleet, where each truck is idle for less time.


We propose a hybrid metaheuristic based algorithm to schedule the crews according to the truck routes determined in a first stage.
This algorithm is capable of exploring solutions where the crews can have 1 or 2 drivers through two problem-specific moves. 
We have performed computational experiments with a set of instances generated with random requests among 15 Argentine cities.
The results have shown the effectiveness of our algorithm, mostly due to the possibility of violating some of the rest constraints during the construction of the initial solutions (which are later repaired) and the option of changing the start time of the tasks.
The proposed repair procedure plays a fundamental role in the search for feasible solutions while the perturbation procedure does so in improving the objective function.
Concerning the main feature of our problem, we can assert that the possibility of carrying an additional driver has led to remarkable cost improvements (about 60\%), in comparison to the usage of individual crews.
In conclusion, the developed algorithm has shown to be effective, and even for the variant \emph{no-shuttle} if more drivers are considered. It is also robust in the sense that it can be adapted to other scenarios with different rest regulations.

As future work, we plan to investigate whether a greater connection between both stages of the sequential approach can produce better solutions.
For instance, by taking into account some requirements of the drivers during the first stage could make the scheduling of the crews easier during the second.
It would also be necessary to weigh the relationship among the objective functions of both stages to globally compare the solutions.


Another line of research is to analyse our problem in the context of the \emph{Physical Internet Initiative} (whose intention is to replace
traditional logistical models to a more sustainable and efficient way by applying ideas of how Internet works), where drivers can be transported from their home to some location and then use another truck to come back to their hometown.

\section*{Appendix}

Here, we present integer programming models for both stages.\\

\noindent \emph{Integer programming model for first stage}. 
Let us consider a simple digraph $G = (U, A)$ with weights $w \in {\mathbb{R}^+_0}^V$ where
$$U = \{u_v : v \in V\} \cup \{u_r : r \in R\} \cup \{u_s\},$$
this is, a node for each truck and request and a sink node $u_s$. The set of arcs $A$ is defined as follows:
\begin{itemize}
    \item There is an arc $(u_v, u_r)$ for each $v \in V$ and each $r \in R$, with weight
    $w(u_v, u_r) = travelDist(l_v, l^p_r) + travelDist(l^p_r, l^d_r)$. It models that $r$ is the first request picked by truck $v$.
    \item There is an arc $(u_{r_1}, u_{r_2})$ for $r_1, r_2 \in R$, with weight
    $w(u_{r_1}, u_{r_2}) = travelDist(l^d_{r_1}, l^p_{r_2}) + travelDist(l^p_{r_2}, l^d_{r_2})$.
    It models that the same truck picks up $r_2$ after it delivers $r_1$.
    \item There is an arc $(u_v, u_s)$ for each $v \in V$ with weight $w(u_v, u_s) = 0$. This arc allows the model to not
    route every truck of the fleet.
    \item There is an arc $(u_r, u_s)$ for each $r \in R$ with weight $w(u_r, u_s) = 0$.
\end{itemize}

A set of truck routes can be represented by means of a binary decision variable $x_e$ for each $e \in A$, where
$x_e = 1$ if and only if some truck travels arc $e$. For each request $r \in R$, we also consider four integer decision variables
$d^p_r$, $d^d_r$, $h^p_r$ and $h^d_r$, which stand for the day or hour the request is picked up or delivered, respectively.
The formulation is given below,
where the notation $e = (\_, s)$ means that the head of the arc $e$ is $s$ regardless of the tail (the same applies for the
notation $(s, \_)$ where the tail is fixed to $s$), and $M$ is a large value, e.g., $M = 24 H + \sum_{e \in A} w(e)$.

\begin{align*}
	& \textrm{Minimise}~~ Z_1 = \sum_{r \in R} c_r (d^d_r - day^d_r) \text{~~~and~~~} Z_2 = \sum_{e \in A} w(e) x_e &
\end{align*}

\noindent subject to:
{ \small
\begin{align}
	&\sum_{e = (\_,u_r) \in A} x_e = 1, &\forall~ r \in R, \label{RESTR1} \\
	&\sum_{e = (u_v,\_) \in A} x_e = 1, &\forall~ v \in V, \label{RESTR2} \\
	&24 d^p_r + h^p_r + 1 + travelTime^T(l^p_r, l^d_r) \leq 24 d^d_r + h^d_r, &\forall~ r \in R, \label{RESTR3} \\
	& travelTime^T(l_v, l^p_r) \leq 24 d^p_r + h^p_r + M (1 - x_{(u_v,u_r)}), &\forall~ v \in V,~ r \in R, \label{RESTR4} \\
	& 24 d^d_{r_1} + h^d_{r_1} + 1 + travelTime^T(l^d_{r_1}, l^p_{r_2})  & \notag \\
	& ~~~~~~~~~~~~~~~~~~ \leq 24 d^p_{r_2} + h^p_{r_2} + M (1 - x_{(u_{r_1},u_{r_2})}), &\forall~ r_1, r_2 \in R,~ r_1 \neq r_2, \label{RESTR5} \\
	&\sum_{e = (\_,u_r) \in A} x_e = \sum_{e = (u_r,\_) \in A} x_e, &\forall~ r \in R, \label{RESTR6} \\
	&x_e \in \{0, 1\}, &\forall~ e \in A, \label{RESTRBOUNDS1} \\
	&d^p_r \in \mathbb{Z} \cap [day^p_r, H),~ d^d_r \in \mathbb{Z} \cap [day^d_r, H), &\forall~ r \in R, \label{RESTRBOUNDS2} \\
	&h^p_r \in \mathbb{Z} \cap [a^p_r, b^p_r],~ h^d_r \in \mathbb{Z} \cap [a^d_r, b^d_r], &\forall~ r \in R. \label{RESTRBOUNDS3}
\end{align} }

Constraints \eqref{RESTR1} establish that all requests are fulfilled.
Constraints \eqref{RESTR2} ensure that all trucks depart from their initial locations.
Constraints \eqref{RESTR3} forbid that a request $r$ could be delivered before it is picked up, taking into account the travel time
between locations $l^p_r$ and $l^d_r$, and the duration of the service of one hour.
Constraints \eqref{RESTR4} and \eqref{RESTR5} consider the time for a truck $v$ to reach the location of its first request $r$ as long as
$x_{(u_v,u_r)} = 1$, and the time for some truck to reach the location of a request $r_2$ after delivering another request $r_1$
as long as $x_{(u_{r_1},u_{r_2})} = 1$, respectively.
Finally, constraints \eqref{RESTR6} are flow conservation equations and \eqref{RESTRBOUNDS1}-\eqref{RESTRBOUNDS3}
establish the domain of the decision variables (integrality and bound constraints).

Note that the elements $T$ and $req$ can be generated from the optimal solution of this model straightforwardly.\\

\noindent \emph{Integer programming model for second stage}. 
We consider the digraph $G = (\mathcal{V}, \tilde{\mathcal{A}})$, having 
$$\tilde{\mathcal{A}} = \{ (s_l, n_t) : l \in L_D, t \in T\} \cup \{ (n_{t_i}, n_{t_j}) : t_i, t_j \in T,~ t_i \neq t_j\} \cup \{ (s_l, \tilde{s}) : l \in L_D \} \cup \{ (n_t, \tilde{s}) : t \in T \}$$
with weights $w_e = shuttleCost(l, p_t)$ for all $e = (s_l, n_t)$,
$w_e = shuttleCost(q_{t_i}, p_{t_j})$ for all $e = (n_{t_i}, n_{t_j})$, and
$w_e = 0$ for all $e = (\_, \tilde{s})$.
Note that $\mathcal{A} \subset \tilde{\mathcal{A}}$ for any
$G_\delta = (\mathcal{V}, \mathcal{A})$, so $\tilde{\mathcal{A}}$ can be seen as a superset of arcs.
Let $\delta(t)$ be an integer variable for each $t \in T$, $x_{de}$ be a binary variable for each $d \in D$ and $e \in \tilde{\mathcal{A}_d} 
\doteq \tilde{\mathcal{A}} \setminus \{(s_l,\_) : l \neq l_d\}$
such that $x_{de} = 1$ if and only if $e$ belongs to the directed path in $G_\delta$ corresponding to the driver $d$,
and $y_e$ be a binary variable for each $e \in \tilde{\mathcal{A}}$ such that $y_e = 1$ if and only if $e$ belongs to some directed path in $G_\delta$.
Then the formulation can be stated as follows.





\begin{align*}
	& \textrm{Minimise}~~ Z_3 = \sum_{d \in D} \sum_{e \in \tilde{\mathcal{A}_d}} w_e x_{de} & 
\end{align*}
subject to:
{ \small
\begin{align}
    & 1 \leq \sum_{d \in D} \sum_{e=(\_,n_t) \in \tilde{\mathcal{A}_d}} x_{de} \leq 2, & \forall~ t \in T, \label{RE1}\\
    & \sum_{e=(s_{l_d}, \_) \in \tilde{\mathcal{A}_d}} x_{de} = \sum_{e=(\_, \tilde{s}) \in \tilde{\mathcal{A}_d}} x_{de} = 1, & \forall~ d \in D, \label{RE2}\\
    & \sum_{e=(\_, n_t) \in \tilde{\mathcal{A}_d}} x_{de} = \sum_{e=(n_t, \_) \in \tilde{\mathcal{A}_d}} x_{de},
          & \forall~ d \in D,~ t \in T, \label{RE3}\\
    & x_{de} \leq y_e, & \forall~ d \in D,~ e \in \tilde{\mathcal{A}_d}, \label{RE4}\\
    & travelTime^S(l_d, p_t) y_{(s_{l_d},n_t)} \leq \delta(t), & \forall~ d \in D,~ t \in T \label{RE5}\\
    & \delta(t_i) + a_{t_i} \leq \delta(t_j), & \forall~ t_i, t_j \in T : \textrm{$t_j$ is next to $t_i$ in some truck route} & \label{RE6}\\
    & \delta(t_i) + a_{t_i} + travelTime^S(q_{t_i}, p_{t_j}) \leq \delta(t_j) + M (1 - y_{(n_{t_i},n_{t_j})}), & \notag \\
    & ~~~~~~~~~~~~~~\forall~ t_i, t_j \in T : \textrm{$t_i$ and $t_j$ belong to different truck routes} & \label{RE7}\\
    & x_{de} \in \{0,1\}, & \forall~ d \in D,~ e \in \tilde{\mathcal{A}_d},  \label{REBOUND2} \\
    & y_e \in \{0,1\}, & \forall~ e \in \tilde{\mathcal{A}},  \label{REBOUND3} \\
    & \delta(t) \in \mathbb{Z}_+, & \forall~ t \in T^t,  \label{REBOUND4} \\
    & \delta(t) \in \bigcup_{i=day_{req(t)}^p}^{H-1} \{24*i+a^p_{req(t)}, \ldots, 24*i+b^p_{req(t)}\}, & \forall~ t \in T^p,  \label{REBOUND5} \\
    & \delta(t) \in \{24*d^d_{req(t)}+a^d_{req(t)}, \ldots,24* d^d_{req(t)}+b^d_{req(t)}\}, & \forall~ t \in T^d.  \label{REBOUND6}
\end{align} }

Constraints \eqref{RE1} ensure that each task is covered by one or two drivers. Constraints \eqref{RE2} set a single initial (and final) arc for each driver according to its initial location. 
Constraints \eqref{RE3} are flow conservation equations.
Constraints \eqref{RE4} control the activation of $y$.
Constraints \eqref{RE5} and \eqref{RE7} enforce Definition~\ref{Def Start Time Assigment}.
Constraints \eqref{RE6}, \eqref{REBOUND5}, and \eqref{REBOUND6} enforce Definition~\ref{defstarttime}.
The remaining constraints define the domain of variables. Note that we set the delivery date of the requests from the optimal solution of the first stage in order to maintain the value of Z1 ($d^d_{req(t)}$ stand for the day the request $req(t)$ is delivered in the optimal solution of the first stage).
We omitted the rest constraints in this formulation for the sake of simplicity. 

\section*{Acknowledgments}

This work was partially supported by grants PICT-2016-0410 and PICT-2017-1826 (ANPCyT), PID ING538 (UNR) and UBACyT 20020170100484BA. We would also like to thank the referees for having read our work carefully and for giving us suggestions that substantially improved it.

\bigskip


\nocite{*}

\bibliographystyle{itor}
\bibliography{itor}

\end{document}